\documentclass[sigconf]{acmart}

\setcopyright{none}
\makeatletter
\def\@ACM@cc@image{}
\def\@ACM@print@copyright{}
\makeatother

\AtBeginDocument{%
  }
    
\usepackage[utf8]{inputenc} 
\usepackage[T1]{fontenc}    
\usepackage{hyperref}       
\usepackage{url}            
\usepackage{amsfonts}       
\usepackage{nicefrac}       
\usepackage{microtype}      
\usepackage{xcolor}         
\usepackage{multirow}
\usepackage{amsmath}
\usepackage{graphicx}
\usepackage{subcaption}
\usepackage{bm}
\usepackage{amsthm}
\usepackage{algorithmic}
\usepackage{algorithm}
\usepackage{float} %
\usepackage{enumitem}    
\usepackage{soul}

\usepackage{makecell}
\usepackage{balance} 
\acmConference[WWW '26] {Proceedings of the ACM Web Conference 2026}{April 13--17, 2026}{Dubai, United Arab Emirates.}
\acmBooktitle{Proceedings of the ACM Web Conference 2026 (WWW '26), April 13--17, 2026, Dubai, United Arab Emirates}
\acmPrice{}
\acmDOI{10.1145/3774904.3792542}
\acmISBN{979-8-4007-2307-0/2026/04}

\begin{document}

\title{Embedding Enhancement via Fine-Tuned Language Models for Learner-Item Cognitive Modeling}

\author{Yuanhao Liu}
\authornote{These authors contribute equally to this research.}
\orcid{0009-0007-3940-6728}
\affiliation{%
  \institution{East China Normal University}
  \city{Shanghai}
  \country{China}
}
\email{51275901044@stu.ecnu.edu.cn}

\author{Zihan Zhou}
\orcid{0009-0009-7784-4846}
\authornotemark[1]
\affiliation{%
  \institution{East China Normal University}
  \city{Shanghai}
  \country{China}
}
\email{zhzhou@stu.ecnu.edu.cn}

\author{Kaiying Wu}
\orcid{0009-0009-7408-7149}
\authornotemark[1]
\affiliation{%
  \institution{East China Normal University}
  \city{Shanghai}
  \country{China}
}
\email{10235102479@stu.ecnu.edu.cn}

\author{Shuo Liu}
\orcid{0000-0001-7970-3187}
\affiliation{%
  \institution{Tencent Inc}
  \city{Shenzhen}
  \country{China}
}
\email{seokliu@tencent.com}

\author{Yiyang Huang}
\orcid{0009-0001-1339-1245}
\affiliation{%
  \institution{East China Normal University}
  \city{Shanghai}
  \country{China}
}
\email{10235102470@stu.ecnu.edu.cn}

\author{Jiajun Guo}
\orcid{0000-0003-4379-2661}
\affiliation{%
  \institution{East China Normal University}
  \city{Shanghai}
  \country{China}
}
\email{jjguo@psy.ecnu.edu.cn}

\author{Aimin Zhou}
\orcid{0000-0002-4768-5946}
\affiliation{%
  \institution{East China Normal University}
  \city{Shanghai}
  \country{China}
}
\affiliation{%
  \institution{Shanghai Innovation Institute}
  \city{Shanghai}
  \country{China}
}
\email{amzhou@cs.ecnu.edu.cn}

\author{Hong Qian}
\orcid{0000-0003-2170-5264}
\authornote{Corresponding author: Hong Qian (hqian@cs.ecnu.edu.cn).}
\affiliation{%
  \institution{East China Normal University}
  \city{Shanghai}
  \country{China}
}
\affiliation{%
  \institution{Shanghai Innovation Institute}
  \city{Shanghai}
  \country{China}
}
\email{hqian@cs.ecnu.edu.cn}

\renewcommand{\shortauthors}{Yuanhao Liu et al.}

\begin{abstract}

Learner-item cognitive modeling plays a central role in the web-based online intelligent education system by enabling cognitive diagnosis (CD), the upstream and crucial component of the system, across increasingly diverse online educational scenarios. Although ID embedding remains the mainstream approach in cognitive modeling due to its effectiveness and flexibility, recent advances in language models (LMs) have introduced new possibilities for incorporating rich semantic representations to enhance CD performance. However, current studies often focus on a specific task, such as zero-shot CD, limiting the broader application of LMs in this field. This highlights the need for a comprehensive analysis of how LMs enhance embeddings through semantic integration across mainstream CD tasks. This paper identifies two key challenges in fully leveraging LMs in existing work: Misalignment between the training objectives of LMs and CD models creates a distribution gap in feature spaces, hindering the potential of LMs for embedding enhancement; A unified framework is essential for integrating textual embeddings across varied CD tasks while preserving the strengths of existing cognitive modeling paradigms, such as ID embeddings, to ensure the robustness of embedding enhancement. To address these challenges, this paper introduces EduEmbed, a unified embedding enhancement framework that leverages fine-tuned LMs to enrich learner-item cognitive modeling across diverse CD tasks. EduEmbed operates in two stages. In the first stage called role-aware interactive fine-tuning, we fine-tune LMs based on role-specific representations and an interaction diagnoser to bridge the semantic gap of CD models. In the second stage called adapter-aware representation integration, we employ a textual adapter to extract task-relevant semantics and integrate them with existing modeling paradigms to improve generalization across diverse CD tasks. We evaluate the proposed framework on four CD tasks and computerized adaptive testing (CAT) task, achieving robust performance. Further analysis reveals the impact of semantic information across diverse tasks, offering key insights for future research on the application of LMs in CD for online intelligent education systems.

\end{abstract}

\maketitle
\newcommand\webconfavailabilityurl{https://doi.org/10.5281/zenodo.18301397}
\ifdefempty{\webconfavailabilityurl}{}{
\begingroup\small\noindent\raggedright\textbf{Resource Availability:}\\
The source code of this paper has been made publicly available at \url{\webconfavailabilityurl} and \url{https://github.com/BW297/EduEmbed}.
\endgroup
}

\begin{figure*}[htbp]
\centering
    \includegraphics[width=0.8\linewidth]{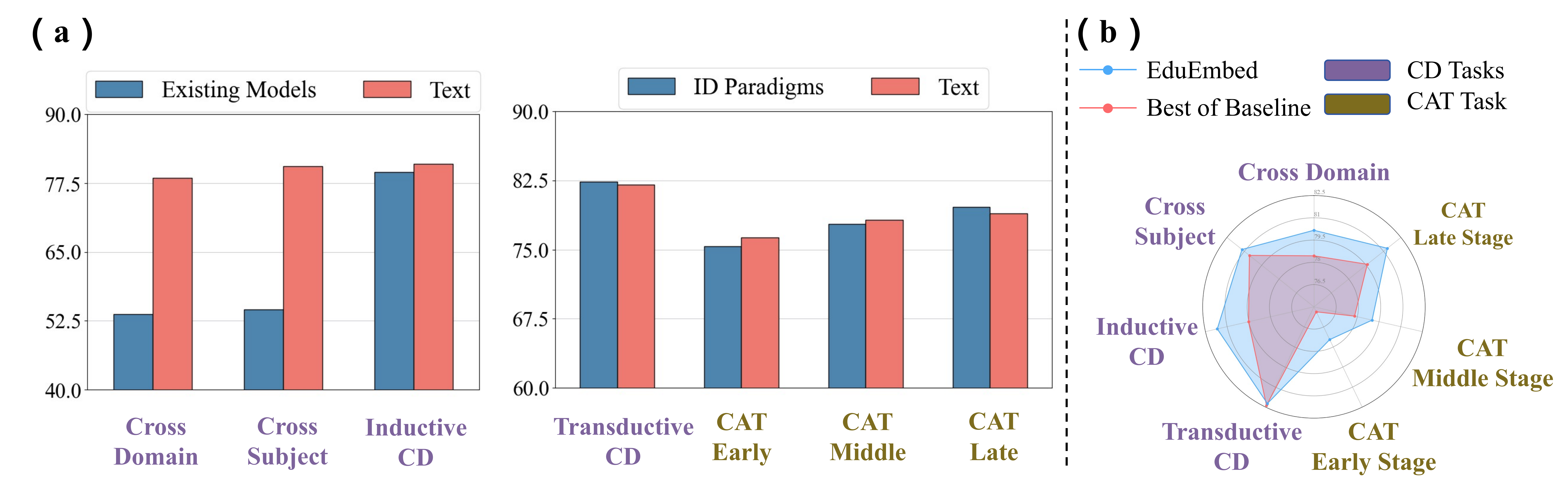}\\
    \caption{(a) Motivation study. (b) The comparison of our proposed EduEmbed with best-performing baseline methods on SLP.}
    \label{fig:intro}
\end{figure*}

\section{Introduction}\label{sec:intro}

With the growing demands of personalized learning, web-based online intelligent education systems~\cite{li2025level} have emerged as a critical development direction. Cognitive Diagnosis (CD)~\cite{Wang2024survey,FineCD,Liu2023New,shen2024symbolic,Liu2023Qccdm}, as a crucial upstream component of the system, aims to infer students' mastery level of specific concepts by analyzing their past interaction records. The diagnosis results can also support further customized applications, such as Computerized Adaptive Testing (CAT)~\cite{ncat,zhuang2023BECAT}. Currently, these technologies have been widely applied in modern web-based online education platforms~\cite{Assist0910}, and single-task scenario settings are no longer sufficient to meet real-world demands. For example, in the field of CD, a variety of scenarios have been proposed and actively studied, including traditional transductive CD~\cite{MIRT, NCDM, KANCD} for daily practice tests, inductive CD~\cite{Liu2024Icdm, li2024towards, liu2025dual} for large-scale, dynamic open student learning environments, zero-shot CD~\cite{techcd, Zero13, LRCD} for interdisciplinary and cross-domain settings and CAT~\cite{ncat,zhuang2023BECAT,Bi2020MAAT}, as a downstream application of CD, for online standardized testing scenarios. 

As the foundational module for CD, learner-item cognitive modeling~\cite{RCD,qian2024orcdf, hiercdf} learns latent representations of learners (e.g., students) and items (e.g., exercises, concepts) via embedding construction, and its quality directly affects aforementioned task performance. ID embedding, which maps entity IDs to latent vectors, has long been the dominant paradigm due to its effectiveness and flexibility, but it struggles to generalize across increasingly diverse CD tasks. Recently, the advancements in language models (LMs)~\cite{devlin2019bert,qwen2.5,Llama} offer new possibilities. Natural language offers a unified interface for modeling diverse CD tasks and pretraining, particularly in large language models, captures rich open-world knowledge, enabling more informative semantic representations. However, most LM-based CD works remain limited to single tasks such as zero-shot CD~\cite{Zero13, LRCD}. \textbf{\emph{Therefore, there is a lack of a comprehensive analysis on the effectiveness of textual semantic embedding generated by LMs across mainstream CD tasks.}}

As shown in Figure~\ref{fig:intro}~(a), we compare the pure textual embeddings generated by the original LMs without any additional training against the best-performing models in each task that do not use textual embeddings, across multiple CD scenarios and different stages of CAT. Detailed experimental settings are provided in Appendix~\ref{apx:motivation}. The results show that the embedding enhancement brought by textual semantic information varies across different CD tasks. Therefore, understanding the enhancement these embeddings bring to each task, as well as the potential improvement space introduced by incorporating textual semantic information in different CD tasks, is essential for assessing the value of textual semantic embedding enhancement and guiding future applications of LMs in CD. In investigating this, we identify two widespread challenges for applying LMs to CD in current research: \textbf{(1) Training objectives misalignment:} A key challenge lies in the misalignment between the training objectives of general LMs and learner-item cognitive modeling in CD models. This often leads to a distribution gap between LM-generated embeddings and the feature space of mainstream CD frameworks, limiting the potential of LMs for embedding enhancement. Aligning LMs semantic pattern with CD models representation may be crucial to unlock full potential of LMs in embedding enhancement. \textbf{(2) Lack of a unified integration framework:} Given the diversity of CD tasks, there is currently no unified integration paradigm that allows textual embeddings to be seamlessly incorporated across varied scenarios while preserving the strengths of existing learning paradigms, such as ID embeddings. This lack of generalizability makes it difficult to ensure a performance lower bound across tasks, thereby limiting the robustness of embedding enhancement.

To address these challenges, this paper proposes EduEmbed, a unified embedding enhancement framework that leverages fine-tuned LMs to enrich learner-item cognitive modeling across diverse CD tasks. The framework consists of two stages. In the first stage, it is assumed that LMs have acquired extensive external knowledge during pretraining. Therefore, we aim to activate their capacity for learner-item cognitive modeling through fine-tuning, which facilitates their adaptation to CD models by aligning the training objectives of LMs with those of CD models to a certain extent. We propose role-aware interactive fine-tuning, where we 
produce textual embeddings aligned with CD models feature spaces, thereby unlocking the full potential of embedding enhancement. In the second stage, adapter-aware representation integration, we 
propose a unified paradigm to integrate mainstream ID embeddings and textual embeddings. By preserving the strengths of ID embeddings, this paradigm enhances the generalization and robustness of embedding enhancement across diverse CD tasks. Benefiting from this two-stage design, EduEmbed consistently achieves robust performance on four representative CD tasks and a downstream CAT task. Moreover, the analysis of the impact of semantic information under diverse CD tasks offers valuable insights for future research about LMs application in CD for online intelligent education systems.

\section{Related Work}

\subsection{Learner-Item Cognitive Modeling in Cognitive Diagnosis}\label{sec:cog_modeling}

CD is a vital field in educational psychology, which is used to infer students' mastery levels for each concept by their response logs. Since responses are noisy indicators influenced by guessing and item properties, a student's mastery level is considered as latent, determining response correctness together with these related properties. Learner-item cognitive modeling serves as the representation learning module in CD, aiming to construct latent representations of learners (e.g., students) and items (e.g., exercises, concepts) via embedding. Most existing methods follow the ID-based embedding paradigm. They can be divided by mastery dimension into two types: latent factor models (e.g., MIRT~\cite{MIRT}) that represent students' mastery as fixed-length vectors, and concept-based models (e.g., DINA~\cite{DINA}) that use concept-specific mastery patterns. With deep learning advancements, more flexible models have emerged. For example, NCDM\cite{NCDM} uses MLPs as interaction functions and models mastery as continuous variables in $[0,1]$. Recent learner-item cognitive modeling methods include MLP-based\cite{KANCD}, graph-based~\cite{RCD, qian2024orcdf}, and Bayesian network-based methods~\cite{hiercdf}.

However, with the increasing diversity of CD task scenarios, the ID-based paradigm is no longer sufficient to support all applications. In inductive CD, IDCD~\cite{li2024towards} replaces ID embeddings with interaction matrices to model the cognitive states of entities. In zero-shot CD, TechCD~\cite{techcd} leverages transferable hand-crafted knowledge graph structures to overcome the limitations of ID embeddings across domains. Meanwhile, models like ZeroCD~\cite{Zero13} and LRCD~\cite{LRCD} introduce textual semantic representation learning to replace ID embeddings, significantly enhancing generalization in zero-shot CD tasks. It is evident that LMs have begun to emerge in learner-item cognitive modeling, but their use in CD remains limited. Given the strong generalization ability of natural language, its potential across diverse CD scenarios deserves deeper exploration.

\subsection{The Appilication of Language Models in Intelligent Education}
Among the major application scenarios for LMs in education, two related scenarios are introduced as follows. 
First, LMs are employed as agents to simulate learner behavior. For example, EduAgent~\cite{EduAgent} leverages LLM-based agents to mimic learners’ engagement with PowerPoint presentations and videos. Agent4Edu~\cite{Agent4Edu} uses LLM as response generators to simulate learner response data, thereby supporting the training and evaluation of downstream educational tasks. Second, LMs have been used as embedders to encode textual information into vector representations, which is the focus of our work. For instance, NCDM+~\cite{NCDM} utilizes exercise text via TextCNN~\cite{Kim2014Textcnn} to complete the Q-Matrix in CD. ECD~\cite{Zhou2021Ecd}, which fuses student context-aware features (e.g., parental education level, monthly study expenses) into representations of students in cognitive diagnosis. ZeroCD~\cite{Zero13} use exercise contents~\cite{nlp1} as textual features to serve as a mediator between the students in source and target domains.  LRCD~\cite{LRCD} further analyzes the behavior patterns among students, exercises, and knowledge concepts to construct unified textual cognitive representations, supporting zero-shot CD. Depite these efforts, current applications of LMs in CD are still simplistic, lacking in-depth adaptation, which may limit their effectiveness. Moreover, most existing methods rely heavily on rich textual data, failing to fully leverage the broad knowledge coverage of LMs and thus, limiting the effectiveness of these methods in real-world educational scenarios.

Although these embedding-based approaches have shown improvements in educational tasks, most of them still rely on LLMs. The lack of deep adaptation to educational datasets often results in suboptimal embeddings, limiting the effectiveness of these methods in real-world educational scenarios.

\section{Preliminaries}

Consider an educational scenario of a web-based online intelligent education system, which involves \(M \) students \(
S=\left\{ s_1,s_2,...,s_M \right\} 
\), \(N\) exercises \(E=\left\{ e_1,e_2,...,e_N \right\} \), and \(K\) concepts \(C=\left\{ c_1,c_2,...,c_K \right\}\). The corresponding response logs \(R
=\{ ( s_i,e_j,r_{ij} ) |s_i\in S,e_j\in E, r_{ij}\in \left\{ 0,1 \right\}\} 
\) consist of a set of triplets \((s_i,e_j,r_{ij})\), where \(r_{ij}\) represents the score obtained by student \(s_i\) on exercise \(e_j\). \(r_{ij}=1\) indicates that the student answered the question correctly and \(r_{ij}=0\) indicates otherwise. Additionally, \(
\bm{Q}=\left\{ q_{j,k} \right\} _{N\times K}
\) is a binary matrix representing the relationship between exercises and concepts, where \(q_{j,k}=1\) indicates that exercise \(e_j\) relates to concept \(c_k\) and \(q_{j,k}=0\) indicates otherwise. 

\textbf{Cognitive Diagnosis Basis}. Given the student's response log \( R \) and the matrix \( \bm{Q} \), the goal of the CD task is to infer the student's mastery \(
\bm{Mas}\in\mathbb{R}^{M\times K}\) on knowledge concepts. Building on this, we will introduce the following four specific educational scenarios and provide detailed explanations of their application in experiments.

$\bullet$ Transductive Cognitive Diagnosis. In this scenario, we assume the set of students and exercises is known and fixed. The CD model uses the known student-exercise score matrix \(\bm{A} \in \mathbb{R}^{M \times N} \) and the exercise-concept relationship matrix \(\bm{Q} \in \mathbb{R}^{N \times K} \) to infer the latent knowledge mastery \( \bm{Mas} \in \mathbb{R}^{M \times K} \) of all students. The goal of this method is to infer students' 
mastery based on the existing response data. 

$\bullet$ Inductive Cognitive Diagnosis. This scenario takes into account the addition of new students and requires the model to evaluate the knowledge mastery of new students without retraining. Given that the set of existing students \( S_o \) and the set of new students \( S_u \) do not overlap, i.e., \( S_o \cap S_u = \emptyset \), the goal is to predict the knowledge mastery of new students \( \bm{Mas}_{u} \in \mathbb{R}^{|S_u| \times K} \) based on the response data of the existing students, thus enabling inductive reasoning of the model.

$\bullet$ Domain-Level Zero-Shot Cognitive Diagnosis. In this scenario, we assume we have response logs from \(H\) source domains \( R_{s} = \{ R_1, R_2, ..., R_{H} \} \). The goal is to train a CD model on the source domains and then infer in the target domain \(T\), where the target domain has no overlap with the source domain in terms of exercises and concepts, i.e., \(E_s \cap E_t = \emptyset, C_s \cap C_t = \emptyset \). In this case, the CD models adapts to the students \(S_t\) in the target domain and predict their knowledge mastery levels \( \bm{Mas}_t \in \mathbb{R}^{M \times K} \).

\begin{figure*}[htbp]
\centering
\includegraphics[width=0.8\linewidth]{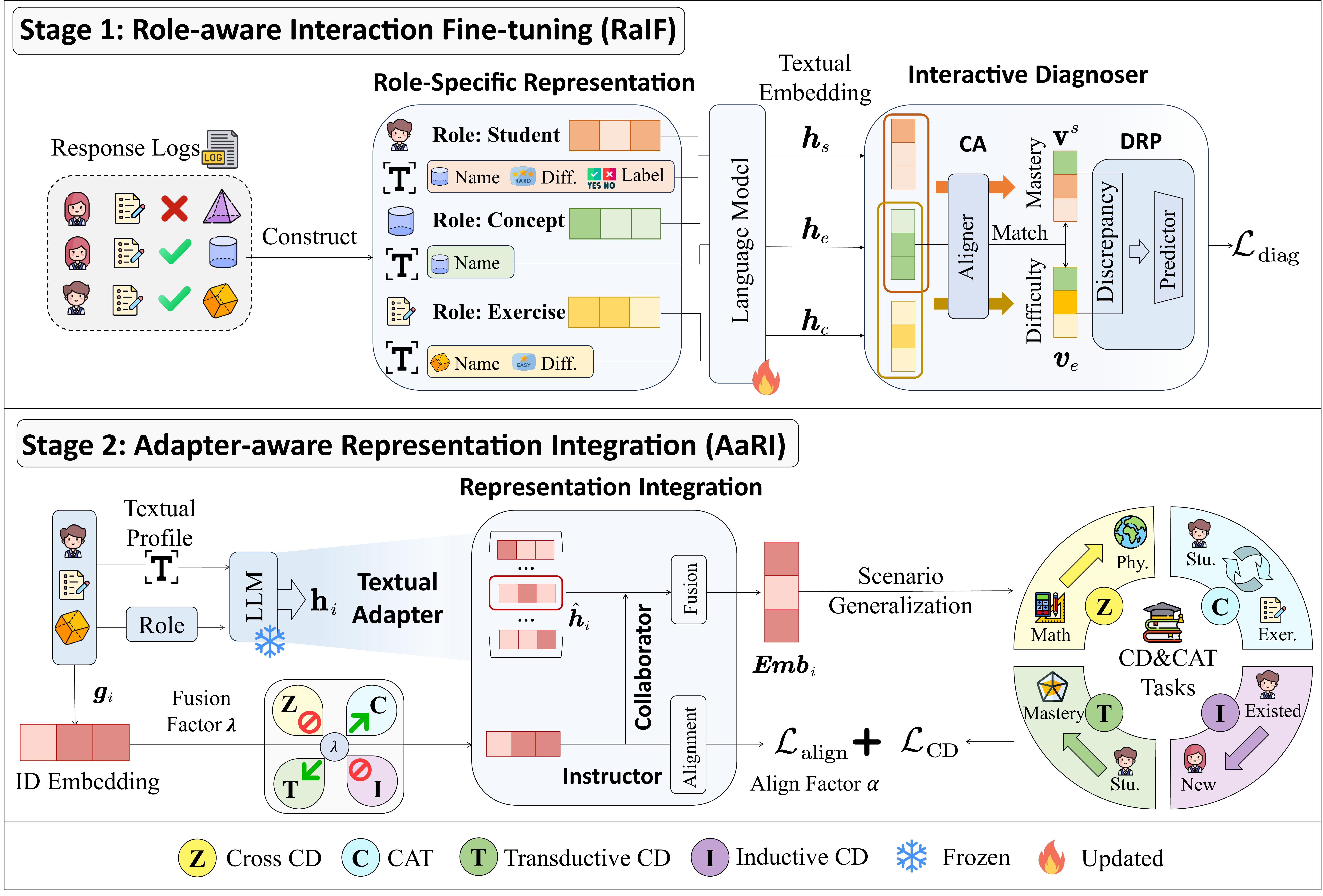}\\
    \caption{The overall framework of the proposed EduEmbed. Stage 1: Role-aware Interaction Fine-tuning (RaIF). Stage 2: Adapter-aware Representation Integration (AaRI).}
    \label{fig:frame}
\end{figure*}

$\bullet$ Computerized Adaptive Testing (CAT). In this scenario, the CD model alternates with the selection strategy to form a feedback loop. At each time step \( t \in [1, T] \), a student \( i \) will update their mastery level based on the answered questions \( R_{t-1,i} = \{ (e_1, r_1), (e_2, r_2), ..., \\(e_{t-1}, r_{t-1}) \} \). The CD models will estimate the student's mastery at time \( t \) as \( \hat{\bm{Mas}}^{t}_i = \bm{Mas}(R_{t-1,i}) \), i.e., the model infers the current mastery level based on previous performance. Then, based on the item selection strategy \( \pi \), the systems will choose a new question \( e_t \) for the student to answer. The student's feedback will update the mastery level. This process will continue for \( T \) steps, with the ultimate goal being for the student's final mastery estimate \( \hat{\bm{Mas}}^T_i \) to be as close as possible to the true ability \( \bm{Mas}^*_i \) at the end of the test.

\textbf{Learner-Item Cognitive Modeling}. Given the response logs $R$ and the Q matrix $\bm{Q}$, the objective of learner-item cognitive modeling is to learn latent representations of learner (e.g., students) and items (e.g., exercises and concepts). These representations for task $t$ are denoted as $\bm{Emb}^t_{s} \in \mathbb{R}^{M \times d_t}$, $\bm{Emb}^t_{e} \in \mathbb{R}^{N \times d_t}$, and $\bm{Emb}^t_{c} \in \mathbb{R}^{K \times d_t}$, respectively, where $d_t$ is the embedding dimension of task $t$. These embeddings serve as foundational representations to support various CD tasks.

\section{Methodology: The proposed EduEmbed}
In this section, we provide a detailed introduction to EduEmbed which consists of two main stages: Role-aware Interaction Fine-tuning and Adapter-aware Representation Enhancement. 
The overall framework of EduEmbed is illustrated in Figure~\ref{fig:frame}.

\subsection{Role-aware Interaction Fine-tuning (RaIF)}\label{sec:RaIF}

This subsection first describes how to design personalized descriptions for three educational roles, students, exercises, and concepts, combined with corresponding encodings to obtain role-specific representations. Then, the constructed textual inputs are fed into the LMs, followed by an explanation of how the model is fine-tuned using an interaction diagnoser to generate textual embeddings that align with CD models.

\subsubsection{Role-specific Representation}\label{sec:role_specific}
\leavevmode\par
Inspired by~\cite{LRCD}, we design personalized descriptions for students, exercises, and concepts to capture their behavior patterns in the dataset. Specifically, the  textual description for each educational role is constructed based on its corresponding attributes $\text{A}$, with the attribute description following a standardized format of $<name \; is\; value>$. Specifically, for concept $c_k$, the attribute is the concept name; for exercise $e_j$, the attributes include the concepts involved and the average accuracy rate $\text{ACR}_{e_j} = \frac{1}{z_j} \sum_{i} r_{ij}$, where $z_j$ denotes the set of students $s$ who have completed exercise \( e_j \) and \( r_{ij} \) denotes the response of student \( s_i \) to exercise \( e_j \); for student $s_i$, the attributes are based on the exercises completed and the corresponding responses. The formal description of attribute $\text{A}$ of the three roles is given below:
\begin{equation}\label{eq:attr}
\left\{
\begin{aligned}
\text{A}_{c_k} &= \text{Name}_{c_k}  \\
\text{A}_{e_j} &= \left[ \left\{ \text{A}_{c_k} \mid \bm{Q}_{j,k} = 1 \right\}, \text{ACR}_{e_j} \right]  \\
\text{A}_{s_i} &= \left\{ \left[ \text{A}_{e_j}, r_{ij} \right] \mid \left( s_i, e_j, r_{ij} \right) \in R \right\}  \\
\end{aligned}.
\right.
\end{equation}

These attributes have minimal dataset demands, making them effective even when textual data is limited. This addresses a key challenge in current educational datasets and enhances real-world applicability. Further analysis on richer textual inputs such as exercise contents is provided in Section~\ref{exp:text}. However, relying solely on descriptions is often insufficient to effectively distinguish educational roles. For example, the textual descriptions of students and exercises may be highly similar, with the only difference being whether there is a response. Such semantic similarity may lead to ambiguity in role alignment within the LMs. Thus, we introduce a token-level learnable role embedding $\bm{p}_\text{role} \in \mathbb{R}^{1 \times d_{\text{LM}}}$ with $\text{role} \in \{\text{Student}, \text{Exercise}, \text{Concept}\}$, which distinguishes three entity types independent of the text descriptions. We define the token combination as follows:
\begin{equation}
\bm{p}=\bm{p}_{\text{base}}+\bm{p}_{\text{role}},
\end{equation}
where $\bm{p}_{\text{base}} \in \mathbb{R}^{1 \times d_{\text{LM}}}$ is the base word token, $\bm{p} \in \mathbb{R}^{1 \times d_{\text{LM}}}$ denotes the final token.
Then we feed $\bm{p}$ into the LMs to obtain the sentence-level textual representation $\bm{h}\in \mathbb{R}^{1\times d}$, where $d$ is the dimension produced by a classification head applied on the LMs' hidden state of the final layer. Notably, as the student $s_i$ may have multiple responses, we apply average pooling to aggregate all corresponding embeddings to obtain the final textual representation $h_{s_i}$. 

\subsubsection{Interactive Diagnoser}
\leavevmode\par
We introduce the interactive diagnoser to fine-tune LMs, thereby aligning the training objectives between LMs and CD models. Thro-\\ugh this design, the textual embeddings generated by the LMs can mitigate the distribution gap in the feature space of CD models to some extent.

\textbf{Concept Aligner.} To enhance the educational interpretability of both students and exercises in the semantic space, we propose a Concept Aligner that projects the textual embeddings of both students and exercises into the concept space. Formally, given the personalized textual embedding of a student $s_i$ as $\bm{h}_{s_i} \in \mathbb{R}^{1 \times d}$ and that of an exercise $e_j$ as $\bm{h}_{e_j} \in \mathbb{R}^{1 \times d}$, we align both to the concept embedding matrix $\bm{H}_c \in \mathbb{R}^{K \times d}$, where $K$ is the number of concepts. We get $\bm{v}_{s_i} = \bm{h}_{s_i} \cdot \bm{H}_c^\top \in \mathbb{R}^{1 \times K}$ as the mastery level of student $s_i$ on each concept $c_k$ and $\bm{v}_{e_j} = \bm{h}_{e_j} \cdot \bm{H}_c^\top \in \mathbb{R}^{1 \times K}$ as the difficulty level of exercise $e_j$ on each concept $c_k$.

\textbf{Discrepancy-based Response Predictor.} Furthermore, we propose a Discrepancy-based Response Predictor (DRP) to model the interaction function between students and exercises. As mentioned in Section~\ref{sec:cog_modeling}, MIRT~\cite{MIRT} is a representative latent factor model that encodes students’ mastery using fixed-dimensional vectors and has been widely used in prior CD studies, where it has consistently shown near-SOTA performance in transductive CD tasks. In this paper, we adopt MIRT as our interaction function to avoid introducing additional learnable parameters during the modeling of student-exercise interactions, which would otherwise require optimizing both the embeddings and the interaction process during fine-tuning, where the predicted score of student $s_i$ on exercise $e_j$ can be formulated as:
\begin{equation}
\hat{r}_{ij} = \sigma(\bm{q}_j^\top (\bm{v}_{s_i} - \bm{v}_{e_j})),
\end{equation}
where $\sigma(\cdot)$ is the sigmoid function and $\bm{q}_j$ denotes the row in the Q matrix $\bm{Q}$ corresponding to exercise $e_j$, indicating the concepts included in exercise $e_j$. Building on this, we apply the BCE loss as the fine-tuning loss for task-specific supervision for interaction modeling. It can be formulated as:
\begin{equation}
\mathcal{L}_{\text{diag}}=-\frac{1}{\left| R \right|}\sum_{\left( s_i,e_j,r_{ij} \right) \in R}{\left[ r_{ij}\log \hat{r}_{ij}+\left( 1-r_{ij} \right) \log \left( 1-\hat{r}_{ij} \right) \right]},
\end{equation}
where $r_{ij} \in \{0, 1\}$ represents the actual response of student $s_i$ to exercise $e_j$ (correct or incorrect) in response logs $R$, and $\hat{r}_{ij}$ is the predicted score.

\subsection{Adapter-aware Representation Integration (AaRI)}\label{sec:AaRI}
This subsection first introduces how to leverage the textual embeddings generated by fine-tuned LMs in Section~\ref{sec:RaIF} by employing a textual adapter to extract task-relevant semantics. Subsequently, we explain how the ID embeddings are utilized to assist in representation integration of the textual embeddings, ultimately producing high-quality embeddings that can be applied to diverse CD tasks.

\subsubsection{Textual Adapter}
\leavevmode\par
We believe that the textual embeddings generated through \textbf{RaIF} in Section~\ref{sec:RaIF} effectively capture general cognitive traits of educational roles. To preserve these general traits, we freeze the fine-tuned LM parameters to ensure consistency across CD tasks. However, since the educational domain involves multiple tasks, each with different demands for these traits, we introduce a textual adapter to extract task-specific semantics. It helps CD models focus on the core traits relevant to the task, thereby significantly enhancing the performance without additional training burdens. The adaptation process can be formulated as:
\begin{equation}
\hat{\bm{h}}^{t}_{s_i}=\mathcal{A}_s^t( \bm{h}_{s_i}\,;\boldsymbol{\theta }_s^t) ,\,\hat{\bm{h}}_{e_j}^{t}=\mathcal{A}_e^t( \bm{h}_{e_j}\,;\boldsymbol{\theta }_e^t) ,\,\hat{\bm{h}}_{c_k}^{t}=\mathcal{A}_c^t( \bm{h}_{c_k}\,;\boldsymbol{\theta }_c^t),
\end{equation}
where $\hat{\bm{h}}^{t}_{s_i},\hat{\bm{h}}_{e_j}^{t},\hat{\bm{h}}_{c_k}^{t}\in \mathbb{R}^{1 \times d_t}$ are the task $t$-relevant embeddings corresponding to student $s_i$, exercise $e_j$, and concept $c_k$, and $d_t$ is the latent dimension in task $t$. $\mathcal{A}_s^t$, $\mathcal{A}_e^t$, and $\mathcal{A}_c^t$ denote the adapters of students, exercises, and concepts for task $t$ respectively, where $\boldsymbol{\theta}_s^t$, $\boldsymbol{\theta}_e^t$, and $\boldsymbol{\theta}_c^t$ are the parameters. In this paper, we represent the adapter as MLPs.
\subsubsection{Representation  Integration}
\leavevmode\par
In this subsection, we propose a unified paradigm for integrating textual and ID embeddings, since ID embeddings serve as a mainstream and effective approach in most CD tasks, particularly in transductive CD~\cite{MIRT,KANCD,qian2024orcdf} and CAT~\cite{ncat,zhuang2023BECAT} task. Specifically, ID embeddings act as both an instructor and a collaborator to guide the alignment and fusion process, aiming to preserve their strengths while ensuring a performance lower bound across various CD tasks.

\textbf{ID Embedding-as-Collaborator.}
To ensure that the final entity embeddings retain rich semantic information while incorporating personalized traits, we introduce the ID embedding $\bm{g}^t$ as a collaborator to the textual embedding $\hat{\bm{h}}^t$ in task $t$. These two representations are jointly fused to produce the latent embedding $\bm{Emb}^t\in \mathbb{R}^{1\times d_t}$, which can be formally expressed as follows:
\begin{equation}
    \bm{Emb}^t=\lambda \cdot\hat{\bm{h}^t}+(1-\lambda)\cdot\bm{g}^t,
\end{equation}
where $\lambda \in [0, 1]$ is the fusion factor that controls the weight of the textual embedding in the fusion of representation. Finally, the learned latent representations are applied to various CD tasks.

\textbf{ID Embedding-as-Instructor.} Since the current textual embeddings are solely derived from learning the behavioral patterns of entities, they may struggle to effectively distinguish between individuals and tend to be sensitive to noisy data. In contrast, ID embeddings often possess stronger discriminative power. Therefore, we introduce ID embeddings as an instructor to align the textual embeddings accordingly, thereby alleviating these limitations. We define our alignment loss based on InfoNCE~\cite{InfoNCE} and take students as an example. We set textual-ID pairs from same students as positive and pairs with other IDs as negative. Specifically,
\begin{equation}
\mathcal{L}_{\text{align},s}^{t} = -\frac{1}{|S|} \sum_{s_i \in S} \log \left( \frac{ 
    \exp( \hat{\boldsymbol{h}}_{s_i}^{t} \cdot \boldsymbol{g}_{s_i}^{t^\top} / \tau )}
{ \sum_{j \ne i} \exp( \hat{\boldsymbol{h}}_{s_i}^{t} \cdot \boldsymbol{g}_{s_j}^{t^\top} / \tau ) } \right),
\end{equation}
where $S$ is the set of students, $\bm{g}_{s_i}^t\in \mathbb{R}^{1\times d_t}$ denotes the ID embeddings for the student $s_i$, and $\tau$ is the temperature hyperparameter. The computation of the alignment loss is similar for exercises and concepts. We obtain the final alignment loss, formulated as $\mathcal{L}_{\text{align}}^t = \mathcal{L}_{\text{align},s}^{t} + \mathcal{L}_{\text{align},e}^{t} + \mathcal{L}_{\text{align},c}^{t}$ for original CD task $t$. Let $\mathcal{L}_{\text{CD}}^t$ denote the loss of task $t$, which is formulated as:
\begin{equation}
\mathcal{L}^t = \mathcal{L}_{\text{CD}}^t + \alpha\cdot \mathcal{L}_\text{align}^t,
\end{equation}
where $\alpha$ is the align factor used to balance the weight of alignment loss $\mathcal{L}_\text{align}^t$.

\section{Experiments}
We conduct experiments on real-world datasets to answer the following key research questions.

$\bullet$ \textbf{RQ1:} How effective is the textual embedding enhancement in EduEmbed across various CD tasks?

$\bullet$ \textbf{RQ2:} How does each component contribute to the performance of EduEmbed across various CD tasks?

$\bullet$ \textbf{RQ3:} How do the types and scale of LMs impact the performance of EduEmbed?

$\bullet$ \textbf{RQ4:} How does the textual attribute selection influence the performance of EduEmbed?

$\bullet$ \textbf{RQ5:} How do hyperparameters influence EduEmbed?

\subsection{Experimental Settings}\label{sec:exp:setting}
\textbf{Datasets Description.} We conduct experiments on four real-world datasets collected from different web-based online intelligent education systems: SLP~\cite{Lu2021slp}, NeurIPS20~\cite{nips2020}, EDM~\cite{edm-cup-2023}, and MOOC~\cite{Yu2023MOOCRadar}. Table~\ref{tab:dataset} provides detailed statistics of those datasets. Here, ``Average Correct Rate'' refers to the mean accuracy of students on exercises, and ``Q Density'' refers to the average number of concepts associated with each exercise. 
Specifically, we implement our Stage 1 RaIF on the SLP-Math dataset, using NeurIPS20 as the in-domain dataset, since both SLP-Math and NeurIPS20 cover junior and senior-level math, and EDM as the out-domain dataset, which focuses on elementary-level math.
This setup allows us to evaluate the generalization performance of EduEmbed across different educational levels. Due to the rich exercise context, MOOC is employed to explore how different attribute selections for textual profiling affect the performance of EduEmbed in RQ5.  All datasets largely satisfy normality due to scale and random splits. The detailed introduction of these datasets is summarized in Appendix~\ref{apx:dataset}. 

\begin{table}[t]
  \centering
  \caption{Statistics of the real-world datasets.}
  \resizebox{0.99\linewidth}{!}{
    \begin{tabular}{l|ccccc}
    \toprule
    \textbf{Datasets} & \textbf{SLP-Math} & \textbf{SLP-Chi} & \textbf{NeurIPS20} & \textbf{EDM} & \textbf{MOOC} \\
    \midrule
    \textbf{\# Students} & 1080 & 562 & 4918   & 2699 & 3000 \\
    \textbf{\# Exercises} & 609 & 510 & 948   & 1479 & 1967 \\
    \textbf{\# Knowledge Concepts} & 32 & 17 & 86 & 319 & 2278 \\
    \textbf{\# Response Logs} & 52100 & 28686 & 1382727   & 116156 & 333602 \\
    \textbf{Average Correct Rate} & 0.506 & 0.623 & 0.545 & 0.628 & 0.812  \\
    \textbf{Q Density} & 1.000 & 1.000 & 4.017   & 1.000 & 2.284 \\
    \bottomrule
    \end{tabular}}%
  \label{tab:dataset}%
\end{table}

\textbf{Evaluation Metrics.} Since students' true mastery levels are unobservable, we follow prior research~\cite{NCDM} to evaluate the performance of EduEmbed by predicting the performance of students on CD tasks. We employ score-prediction metrics and interpretability metrics to assess its effectiveness. Specifically, for score prediction metrics, given that the CD task is a binary classification problem, we use the Area Under the Curve (AUC) and Accuracy (ACC) as evaluation metrics. For interpretability, following previous works~\cite{NCDM}, we employ the Degree of Agreement (DOA) to assess the interpretability of the mastery levels of students. For a more detailed explanation of DOA, please refer to Appendix~\ref{apx:doa}.

\textbf{Compared Methods.} The following provides a brief description of the baselines used in four representative CD tasks and a downstream CAT task.

$\bullet$ \textbf{Transductive CD.} As the most traditional task setting, Transductive CD has been extensively studied, with most methods adopting the ID embedding paradigm, which fits well within our framework. We select three representative models as both compared methods and integrated CD models in EduEmbed: the classic MIRT~\cite{MIRT}, the widely used KaNCD~\cite{KANCD}, and the recent SOTA model ORCDF~\cite{qian2024orcdf}.

$\bullet$ \textbf{Inductive CD.} In inductive CD, traditional ID embedding paradigm is no longer applicable. Therefore, EduEmbed relies solely on textual semantic features in this setting. We compare our approach with two recent models, IDCD~\cite{li2024towards} and ICDM~\cite{Liu2024Icdm}.

$\bullet$ \textbf{Zero-shot CD.} Zero-shot CD can be further divided into two categories. The first is cross-subject CD, which focuses on transfer across different academic subjects, and the second is cross-CD, which addresses transfer across different datasets. In both tasks, the dominant paradigm is textual semantic embeddings. Accordingly, EduEmbed adopts pure textual semantic features in this setting. We compare our approach with three representative methods: TechCD~\cite{techcd}, ZeroCD~\cite{Zero13}, and LRCD~\cite{LRCD}. 

$\bullet$ \textbf{Computerized Adaptive Testing (CAT).} CAT is a downstream task of CD. It consists of two main components: a selection strategy and a CD model. We select NCD~\cite{NCDM} and IRT~\cite{IRTA} as the CD models and five selection strategies: RANDOM, MAAT~\cite{Bi2020MAAT}, BOBCAT~\cite{Aritra2021BOBOCAT}, NCAT~\cite{ncat} and BECAT~\cite{zhuang2023BECAT}. Since CAT follows the ID embedding paradigm, we also integrate ID embeddings into our EduEmbed. 

\textbf{Implementation Details.} 
For stage 1, we use Qwen2.5-3B~\cite{qwen2.5} as the default LM. Large LMs are fine-tuned with LoRA~\cite{Lora}, whereas smaller models undergo full fine-tuning. For stage 2, we set $d_t$ to 64, which is the dimension of the learned latent representations in all tasks. The batch size is set to 256 for all CD tasks, and for CAT task, the batch size is chosen from the set $\{32,64,128,256\}$. The learning rate is chosen from $\{1e^{-4},\,5e^{-4},\,1e^{-3},\,5e^{-3},\,1e^{-2}\}$. All experiments are conducted on two A6000 GPUs. We employ a grid search on the validation set to obtain the best hyperparameters and the detailed hyperparameter analysis is provided in Appendix~\ref{apx:hyper}.


\subsection{Experimental Results}\label{sec:exp:results}

\subsubsection{Effectiveness Analysis of Embedding Enhancement (To RQ1)}\label{sec:emb_enhance}

\begin{table*}[!t]
  \centering
  \caption{The overall performance of EduEmbed compared with the baseline methods in four CD tasks. Within each method, the highest mean performance is highlighted in bold. The value following ``$\pm$'' denotes the standard deviation of the model’s performance. If a mean value is significantly higher than the second-best result according to a $t$-test with a significance level of 0.05, it is marked with ``*''.}
   \resizebox{0.99\linewidth}{!}{
    \begin{tabular}{c|c|ccc|ccc|ccc}
    \toprule
    \multicolumn{2}{c|}{\textbf{Datasets}} & \multicolumn{3}{c|}{\textbf{SLP-Math}} & \multicolumn{3}{c|}{\textbf{NeurIPS20}} & \multicolumn{3}{c}{\textbf{EDM}} \\
    \midrule
    \textbf{Scenarios} & \textbf{Method} & \textbf{AUC} & \textbf{ACC} & \textbf{DOA} & \textbf{AUC} & \textbf{ACC} & \textbf{DOA} & \textbf{AUC} & \textbf{ACC} & \textbf{DOA} \\
    \midrule
    \multirow{4}[1]{*}{\textbf{Transductive CD}} 
    & \textbf{MIRT} & 82.03$\pm$0.01 & \textbf{74.81}$\pm$0.09 & -- & \underline{78.68$\pm$0.01} & \underline{71.77$\pm$0.02} & -- & 78.98$\pm$0.03 & 74.36$\pm$0.04 & -- \\
    & \textbf{KaNCD} & 82.12$\pm$0.13 & \underline{74.67$\pm$0.11} & 77.81$\pm$0.13 & 78.57$\pm$0.03 & 71.73$\pm$0.04 & 66.61$\pm$1.92 & 79.92$\pm$0.13 & 74.40$\pm$0.23 & \textbf{78.78}$^*$$\pm$0.12 \\
    & \textbf{ORCDF} & \textbf{82.37}$\pm$0.01 & 74.48$\pm$0.13 & \textbf{78.24}$^*$$\pm$0.08 & \textbf{78.70}$\pm$0.03 & \textbf{71.79}$\pm$0.03 & \underline{73.58$\pm$0.04} & \textbf{82.63}$\pm$0.07 & \textbf{76.88}$\pm$0.03 & \underline{77.84$\pm$0.16} \\
    & \textbf{EduEmbed} & \underline{82.23$\pm$0.05} & 74.45$\pm$0.11 & \underline{77.85$\pm$0.09} & 78.55$\pm$0.01 & 71.75$\pm$0.02 & \textbf{73.60}$\pm$0.01& \underline{82.59$\pm$0.05} & \underline{76.75$\pm$0.02} & 77.65$\pm$0.11 \\
    \midrule
    \multirow{3}[2]{*}{\textbf{Inductive CD}}
    & \textbf{ICDM}  & 74.54$\pm$0.03 & 68.83$\pm$0.01 & 60.49$\pm$0.02 & 71.72$\pm$0.00 & 65.63$\pm$0.01 & 59.00$\pm$0.00 & 74.18$\pm$0.01 & 70.54$\pm$0.01 &65.38$\pm$0.01\\
    & \textbf{IDCD}  & \underline{79.52$\pm$0.06} & \underline{72.59$\pm$0.12} & \textbf{80.96}$^*$$\pm$0.04 & \underline{75.91$\pm$0.23} & \underline{69.84$\pm$0.20} & \textbf{73.16}$^*$$\pm$0.38 & \underline{79.67$\pm$0.07}& \textbf{75.41}$\pm$0.13 & \textbf{79.93}$^*$$\pm$0.49 \\
    & \textbf{EduEmbed} & \textbf{81.68}$^*$$\pm$0.04 & \textbf{73.78}$^*$$\pm$0.11 & \underline{78.61$\pm$0.05} & \textbf{76.59}$^*$$\pm$0.07 & \textbf{70.01}$\pm$0.17 & \underline{72.78$\pm$0.32} & \textbf{80.66}$^*$$\pm$0.04 & \underline{75.35$\pm$0.44} & \underline{76.53$\pm$0.03} \\
    \midrule
    \multirow{3}[1]{*}{\textbf{Cross-Domain CD}} 
    & \textbf{TechCD}  & 52.52$\pm$0.14 & 53.27$\pm$0.41 & 54.03$\pm$1.16 & 52.05$\pm$0.08 & 53.65$\pm$0.27 & 52.89$\pm$0.71 & 54.05$\pm$0.21 & 63.67$\pm$0.83 & 58.71$\pm$0.43 \\
    & \textbf{LRCD}  & \underline{79.67$\pm$0.69} & \underline{72.11$\pm$0.33} & \underline{76.15$\pm$0.42} & \underline{76.05$\pm$0.31} & \underline{68.47$\pm$1.03}& \underline{73.00$\pm$0.03} & \textbf{79.19}$^*$$\pm$0.21 & \underline{73.02$\pm$1.77} & \underline{76.91$\pm$0.10} \\
    & \textbf{EduEmbed} & \textbf{80.06}$^*$$\pm$0.38 & \textbf{72.61}$^*$$\pm$0.23 & \textbf{78.61}$^*$$\pm$0.14 & \textbf{76.31}$\pm$0.16 & \textbf{69.41}$^*$$\pm$0.43 & \textbf{73.02}$\pm$0.03 & \underline{78.28$\pm$1.13} & \textbf{74.68}$^*$$\pm$0.00 &\textbf{76.95}$\pm$0.00 \\
    \bottomrule
    \end{tabular}
    }
  \label{tab:CD}%
\end{table*}

\begin{table}[!t]
\centering
\captionsetup{type=table}
\caption{The performance of cross-subject CD on SLP. Other details are as same as Table ~\ref{tab:CD}.}
\resizebox{0.9\linewidth}{!}{
\begin{tabular}{@{}c|c|c|c@{}}
\toprule
\textbf{Method}   & \textbf{AUC} & \textbf{ACC} & \textbf{DOA} \\ \midrule
\textbf{LRCD}       & 80.56$\pm$0.12       & 72.59$\pm$0.32       & 76.87$\pm$0.04       \\
\textbf{EduEmbed}   & \textbf{81.20}$^*$$\pm$0.21 & \textbf{73.69}$^*$$\pm$0.42 & \textbf{77.11}$^*$$\pm$0.08 \\ 
\bottomrule
\end{tabular}
}
\label{tab:subject}
\end{table}

\begin{table}[!t]
  \centering
   \caption{The overall performance of EduEmbed with five CAT selection strategies on SLP-Math. ``OL'' stands for the original method under ID embedding paradigm.}
  \resizebox{0.99\linewidth}{!}{
    \begin{tabular}{c|c|cc|cc}
    \toprule
    \multicolumn{2}{c|}{\textbf{Dataset}} & \multicolumn{4}{c}{\textbf{SLP-Math}} \\
    \midrule
    \multicolumn{2}{c|}{\textbf{Metric}} & \multicolumn{4}{c}{\textbf{AUC / ACC (\%)}} \\
    \midrule
    \multirow{2}[1]{*}{\textbf{Strategy}} & \multirow{2}[1]{*}{\textbf{step}} & \multicolumn{2}{c|}{\textbf{IRT}} & \multicolumn{2}{c}{\textbf{NCD}} \\
          &       & \textbf{OL}    & \textbf{EduEmbed} & \textbf{OL}    & \textbf{EduEmbed} \\
    \midrule
    \multirow{3}[1]{*}{\textbf{RANDOM}} & \textbf{5}     & 74.61 / 68.03 & \textbf{75.23}$^{*}$ / \textbf{69.42}$^{*}$ & 73.38 / 67.38 & \textbf{74.01}$^{*}$ / \textbf{68.02}$^{*}$ \\
          & \textbf{10}    & 77.15 / 70.16 & \textbf{78.56}$^{*}$ / \textbf{71.48}$^{*}$ & 76.47 / 69.59 & \textbf{78.20}$^{*}$ / \textbf{71.22}$^{*}$ \\
          & \textbf{15}    & 78.44 / 71.34 & \textbf{80.24}$^{*}$ / \textbf{72.02}$^{*}$ & 78.33 / 70.78 & \textbf{79.28}$^{*}$ / \textbf{72.09}$^{*}$ \\
    \midrule
    \multirow{3}[2]{*}{\textbf{MAAT}} & \textbf{5}     & 74.18 / 67.35 & \textbf{76.66}$^{*}$ / \textbf{69.85}$^{*}$ & 73.66 / 60.07 & \textbf{74.02}$^{*}$ / \textbf{60.82}$^{*}$ \\
          & \textbf{10}    & 76.26 / 68.35 & \textbf{78.96}$^{*}$ / \textbf{71.17}$^{*}$ & 76.29 / 60.77 & \textbf{77.32}$^{*}$ / \textbf{61.23}$^{*}$ \\
          & \textbf{15}    & 77.32 / 69.30 & \textbf{79.42}$^{*}$ / \textbf{71.55}$^{*}$ & 77.88 / 63.65 & \textbf{77.92}$^{*}$ / \textbf{64.21}$^{*}$ \\
    \midrule
    \multirow{3}[2]{*}{\textbf{BOBCAT}} & \textbf{5}     & 75.67 / 68.75 & \textbf{78.95}$^{*}$ / \textbf{71.91}$^{*}$ & 73.74 / 66.39 & \textbf{74.52}$^{*}$ / \textbf{68.35}$^{*}$ \\
          & \textbf{10}    & 77.75 / 70.75 & \textbf{80.44}$^{*}$ / \textbf{72.27}$^{*}$ & 75.69 / 69.05 & \textbf{76.27}$^{*}$ / \textbf{70.14}$^{*}$ \\
          & \textbf{15}    & 78.89 / 71.65 & \textbf{81.07}$^{*}$ / \textbf{73.54}$^{*}$ & 77.43 / 70.57 & \textbf{77.44}$^{*}$ / \textbf{71.05}$^{*}$ \\
    \midrule
    \multirow{3}[1]{*}{\textbf{NCAT}} & \textbf{5}     & 73.94 / 67.35 & \textbf{77.63}$^{*}$ / \textbf{70.30}$^{*}$ & \textbf{73.32} / 62.78 & 73.19 / \textbf{67.08}$^{*}$ \\
          & \textbf{10}    & 75.89 / 68.86 & \textbf{80.14}$^{*}$ / \textbf{72.54}$^{*}$ & 76.30 / 68.71 & \textbf{76.59}$^{*}$ / \textbf{70.03}$^{*}$ \\
          & \textbf{15}    & 77.45 / 70.21 & \textbf{80.43}$^{*}$ / \textbf{72.57}$^{*}$ & 77.43 / 70.67 & \textbf{79.41}$^{*}$ / \textbf{72.09}$^{*}$ \\
    \midrule
    \multirow{3}[1]{*}{\textbf{BECAT}} & \textbf{5}     & 75.37 / 68.76 & \textbf{77.45}$^{*}$ / \textbf{70.40}$^{*}$ & 71.85 / 64.70 & \textbf{72.36}$^{*}$ / \textbf{65.74}$^{*}$ \\
          & \textbf{10}    & 77.81 / 70.95 & \textbf{79.02}$^{*}$ / \textbf{71.48}$^{*}$ & 75.16 / 66.26 & \textbf{77.26}$^{*}$ / \textbf{69.57}$^{*}$ \\
          & \textbf{15}    & 79.60 / 72.70 & \textbf{81.33}$^{*}$ / \textbf{73.38}$^{*}$ & 77.21 / 69.73 & \textbf{78.40}$^{*}$ / \textbf{70.20}$^{*}$ \\
    \bottomrule
\end{tabular}

    }
  \label{tab:cat}
\end{table}

As shown in Table~\ref{tab:CD} and
\ref{tab:subject}, we conduct a detailed analysis of the effectiveness of textual embedding enhancement across different CD tasks. For CAT, the experimental results on SLP-Math dataset in Table~\ref{tab:cat} are shown as an instance. For zero-shot CD, we adopt both cross-subject and cross-domain settings. In the cross-subject CD, we illustrate a representative case where the source domain is the Chinese literature subject (SLP-Chi) and the target domain is the mathematics subject (SLP-Math) within the diverse SLP dataset. For cross-domain CD, for SLP-Math, we use EDM as the source domain and SLP-Math itself as the target domain. Additionally, for in-domain and out-of-domain datasets, we treat each dataset itself as the source domain, with the other dataset serving as the target domain. The complete analysis are provided in Appendix~\ref{apx:tab:CD&CAT}.

\textbf{Significant Enhancement in Cold Start and High Generalization Scenarios.} Textual embedding shows clear performance enhancement in scenarios requiring strong generalization or having severe cold-start issues, such as inductive CD, zero-shot CD and the early stages of CAT.

\textbf{Limited Enhancement in Low Generalization Requirements Tasks.} 
In tasks with low generalization demands, such as transductive CD, textual semantic embedding offers limited enhancement. Therefore, EduEmbed effectively integrates the ID paradigm, ensuring the performance lower bound and maintaining competitive results.

\textbf{Interpretability Analysis.} For models relying entirely on textual semantic features like LRCD, the fine-tuned EduEmbed offers better interpretability. However, for pattern-driven models like IDCD, which use sparse handcrafted interaction features, these features often show clearer structure and thus outperform dense textual embeddings.

\textbf{Domain-Sensitive Enhancement.}
The enhancement provided by fine-tuned LMs is sensitive to their training datasets. As our LM is fine-tuned on SLP-Math, it shows strong performance in in-domain datasets like NeurIPS20, but their generalization to out-domain datasets like EDM remains limited and requires further exploration.

Limited cases like low generalization and out-of-domain applications are discussed in Appendix~\ref{apx:discuss}.

\subsubsection{Ablation Study (To RQ2)}

\begin{table}[!t]
\centering
\caption{Ablation study in four CD tasks on SLP-Math.}
\resizebox{0.99\linewidth}{!}{
\begin{tabular}{c|c|cccc}
    \toprule
    \textbf{CD Scenario} & \textbf{Metric} & \textbf{\makecell{EduEmbed\\w/o RaIF}} & \textbf{\makecell{EduEmbed\\w/o RsR}} & \textbf{\makecell{EduEmbed\\w/o TA}} & \textbf{\makecell{EduEmbed}} \\
    \midrule
    \multirow{3}{*}{\textbf{\makecell{Transductive\\CD}}} 
          & \textbf{AUC}   & \textbf{82.27} & 82.24 & 82.06 & 82.23 \\
          & \textbf{ACC}   & 74.40 & 74.40 & 74.38 & \textbf{74.45} \\
          & \textbf{DOA}   & 77.75 & 77.44 & 76.78 & \textbf{77.85} \\
    \midrule
    \multirow{3}{*}{\textbf{\makecell{Inductive\\CD}}} 
          & \textbf{AUC}   & 81.04 & 81.59 & 81.62 & \textbf{81.68} \\
          & \textbf{ACC}   & 73.75 & 73.63 & \textbf{73.97} & 73.78 \\
          & \textbf{DOA}   & 78.60 & 77.33 & \textbf{78.79} & 78.61 \\
    \midrule
    \multirow{3}{*}{\textbf{\makecell{Cross\\Domain CD}}} 
          & \textbf{AUC}   & 78.49 & 79.87 & 77.45 & \textbf{80.06} \\
          & \textbf{ACC}   & 71.24 & 71.12 & 64.05 & \textbf{72.61} \\
          & \textbf{DOA}   & 76.87 & \textbf{78.91} & 76.22 & 78.61 \\
    \midrule
    \multirow{3}{*}{\textbf{\makecell{Cross\\Subject CD}}} 
          & \textbf{AUC}   & 80.41 & 81.14 & 78.01 & \textbf{81.20} \\
          & \textbf{ACC}   & 72.87 & 73.64 & 63.51 & \textbf{73.69} \\
          & \textbf{DOA}   & 77.01 & \textbf{77.19} & 76.12 & 77.11 \\
    \bottomrule
\end{tabular}
}
\label{tab:ablation_CD}
\end{table}

To validate the efficacy of each module in EduEmbed, we conduct an ablation study. Five ablated versions of EduEmbed are presented. \emph{EduEmbed-w/o-RaIF} omits all the fine-tuning designs, using the textual embeddings generated directly by LMs; \emph{EduEmbed-w/o-RsR} removes the role embedding $\bm{r}_{\text{role}}$ from fine-tuning process; \emph{EduEmbed-w/o-TA} skips the Textual Adapter which is MLPs in this paper; \emph{EduEmbed-w/o-IDI} does not utilize the alignment loss in AaRI; In \emph{EduEmbed-w/o-IDC}, ID embeddings are not integrated with textual embeddings. Specially, \emph{EduEmbed-w/o-TA} replaces MLPs with a simple linear layer in inductive CD and CAT. Also, some ablation experiments cannot be conducted in certain scenarios due to limitations. Corresponding explanations would be given in Appendix~\ref{apx:ablation}.

\textbf{Experimental Results.} As shown in Table~\ref{tab:ablation_CD}, our proposed EduEmbed outperforms most of its ablated versions, confirming the effectiveness of each module. However, we also observe that certain ablated versions exhibit superior performance in specific scenarios. In transductive CD, due to the relatively low requirement for generalization, the performance gains brought by fine-tuning are limited. In inductive CD, using a simple linear layer as the adapter in \emph{EduEmbed-w/o-TA} helps mitigate potential overfitting and achieve strong predictive performance. In zero-shot CD, where a greater generalization of semantics is required, the lack of explicit semantic information in role embeddings limits the interpretability of EduEmbed compared with \emph{EduEmbed-w/o-RsR}. For more results and further analysis, please refer to Appendix~\ref{apx:ablation}.

\subsubsection{Comparison of Types and Scales of the LMs (To RQ3)}\label{exp:scale&type}
\begin{figure}[!t]
\centering
\includegraphics[width=0.99\linewidth]{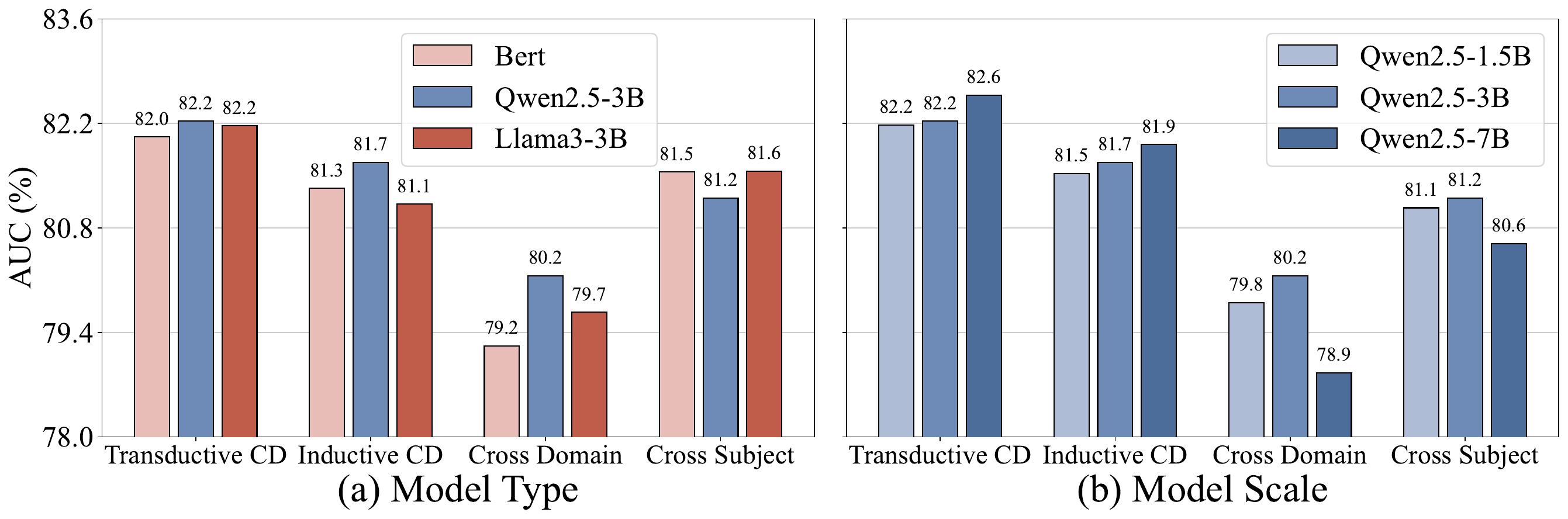}\\
    \caption{The performance of EduEmbed under varying LMs types and scales on SLP-Math.}
    \label{fig:exp:type_and_scale_AUC}

\end{figure}

Here, we investigate the impact of LMs scales and types on the performance of EduEmbed. We conduct experiments on four CD tasks, and the corresponding results based on AUC are shown in Figure~\ref{fig:exp:type_and_scale_AUC}. For more detailed evaluations, please refer to Figure~\ref{fig:exp:type} and ~\ref{fig:exp:scale} in Appendix~\ref{apx:scale&type}.

\textbf{Model Types}. We fine-tune Qwen2.5-3B~\cite{qwen2.5}, Llama3.2-3B~\cite{Llama} and Bert-Base-Cased~\cite{devlin2019bert}, respectively. As shown in Figure~\ref{fig:exp:type_and_scale_AUC}~(a), Qwen2.5-3B delivers optimal performance in most CD scenarios, likely due to its advanced text comprehension and generation capabilities. However, its performance in cross-subject CD is less satisfactory, possibly because it tends to memorize subject-specific patterns from the training data, leading to a limited capacity to generalize to unseen subjects.

\textbf{Model Scales}. We fine-tune the Qwen2.5-series~\cite{qwen2.5} LMs with 1.5B, 3B, and 7B parameters, respectively. 

As shown in the results of Figure~\ref{fig:exp:type_and_scale_AUC} (b), we observe that in transductive CD and inductive CD, model performance improves as the parameter size increases. This is likely due to the similar distribution between training and testing data, which allows larger models to more effectively capture complex cognitive patterns during fine-tuning. However, in cross-domain and cross-subject CD, the performance initially improves but then declines as the model size increases. This trend may be attributed to domain bias in the training data. Larger models tend to overfit fine-grained, domain-specific features, improving in-domain learning but impairing generalization to new domains.

\subsubsection{The Effect of Text Selection (To RQ4)}\label{exp:text}

Previous research~\cite{nlp1} has shown that the textual content of exercises can serve as valuable attributes for learner-item cognitive modeling. However, many existing datasets lack such content, limiting the broader application of text-based features in CD. To assess the impact of this limitation, we conduct experiments on MOOC dataset which includes exercise content, under both inductive CD and transductive CD. Corresponding details are presented in Appendix~\ref{apx:text_selection}. 

\subsubsection{Hyperparameter Analysis (To RQ5)}\label{exp:hyper}

We investigate the impact of two key hyperparameters on the performance of EduEmbed. For detailed results, please refer to Appendix~\ref{apx:hyper}.

\section{Conclusion and Discussion}
In this paper, we systematically evaluate and reveal the task-based potential of LM-based textual embeddings across mainstream CD tasks for web-based online intelligent education systems. We introduce EduEmbed, a unified enhancement framework that leverages fine-tuned LMs to improve learner-item cognitive modeling.
Comprehensive experiments verify the varying enhancement brought by semantic information, offering insights for future research. Limitations and future directions including performance robustness in low-generalization scenarios, further unified integration and computational cost are discussed in Appendix~\ref{apx:discuss}.


\begin{acks}
We would like to thank the anonymous reviewers for their constructive comments. The algorithms and datasets in the paper do not involve any ethical issue. This work is supported by the National Natural Science Foundation of China (No. 62476091), and Tencent Inc Research Program.


\end{acks}
\clearpage
\bibliographystyle{ACM-Reference-Format}
\balance
\bibliography{reference}
\clearpage
\appendix

\section*{Appendix}




\section{Details of Motivation Study}\label{apx:motivation}
In this section, we provide the corresponding settings of our motivation study in Figure~\ref{fig:intro}~(a) presented in Section~\ref{sec:intro}. 

In four CD scenarios and CAT task, we incorporate personalized textual descriptions of students  proposed in Eq.~\ref{eq:attr} as textual embedding features for modeling. In zero-shot CD, this textual embedding model refers to LRCD. For zero-shot CD and inductive CD, we introduce existing models, TechCD and IDCD, respectively, as non-text embedding baselines. In transductive CD and CAT, mainstream ID embeddings are used as non-text embedding baselines. We use IRT as the CD model in CAT. All the results are reported based on AUC.

\section{Experiments}\label{apx:exp}
\subsection{Details about the Datasets}\label{apx:dataset}

In this subsection, we provide detailed introduction of the datasets and the corresponding processing details.

\textbf{Source}. Here we provide the dataset source we use in this paper:

$\bullet$ \textbf{SLP}~\cite{Lu2021slp}: SLP is a K-12 dataset from the online education platform SLP, recording students’ performance across eight subjects over three years (7th to 9th grade). In our paper, we use two subjects: Math and Chinese.

$\bullet$ \textbf{NeurIPS20}~\cite{nips2020}: NeurIPS20 comes from the NeurIPS 2020 Education Challenge, containing student response logs to Eedi math problems over two school years (2018–2020). Eedi is a widely used online learning platform that provides diagnostic multiple-choice questions for middle and high school students.

$\bullet$ \textbf{EDM}~\cite{edm-cup-2023}: Derived from the EDM Cup 2023, EDM captures millions of student interactions on ASSISTments, a web-based K-12 math learning system, with concepts mainly at the elementary level.

$\bullet$ \textbf{MOOC}~\cite{Yu2023MOOCRadar}: Collected from a large-scale Chinese MOOC platform, MOOC offers rich learning resources, fine-grained concepts, behavioral logs, and contextual information such as textual descriptions and annotations.

\textbf{Process}. To ensure sufficient response data, we exclude students with fewer than 10, 10, 30, and 30 responses in SLP, MOOC, NeurIPS20, and EDM, respectively. To reduce computational cost, we randomly sample 3000 students from MOOC. Response logs are split into 70\%/10\%/20\% for training, validation, and testing in both stages. During Stage 1, we cap each student at 50 responses, randomly sampling when necessary. In inductive CD, students are split into existing ($S_o$) and new ($S_u$) groups at a 1:1 ratio, while in CAT, 30\% of responses are used for model pre-training. To prevent information leakage, target-domain test data are excluded from training in zero-shot CD, and student textual embeddings are omitted in CAT.

\subsection{Degree of Agreement (DOA)}\label{apx:doa}
We provide a detailed formulation of the \emph{Degree of Agreement (DOA)} to quantify the alignment between predicted mastery and actual performance. Let $\bm{Mas} \in \mathbb{R}^{M \times K}$ denote the predicted mastery matrix for $M$ students and $K$ concepts. The core intuition is that if student $s_a$ achieves higher accuracy than $s_b$ on exercises of concept $c_k$, then $s_a$ should exhibit greater mastery, i.e., $\bm{Mas}{s_a, c_k} > \bm{Mas}{s_b, c_k}$. The DOA for concept $c_k$ is computed accordingly.

\begin{table}[h]
\centering
\captionsetup{type=table}
\caption{The performance of EduEmbed with overlapping students. Other details are as same as Table ~\ref{tab:CD}.}
\resizebox{0.6\linewidth}{!}{
\begin{tabular}{@{}c|ccc@{}}
\toprule
\textbf{Metric}   & \textbf{AUC}            & \textbf{ACC}            & \textbf{DOA}           \\ \midrule
\textbf{TechCD}   & 57.96          & 56.44          & 48.8          \\
\textbf{ZeroCD}   & 61.77          & 59.07          & 50.81         \\
\textbf{LRCD}     & \underline{78.56}    & \underline{72.01}    & \underline{74.96}   \\
\textbf{EduEmbed} & \textbf{78.74}$^*$ & \textbf{72.32}$^*$ & \textbf{75.30}$^*$ \\ \bottomrule
\end{tabular}
}
\label{tab:Zero_overlap}
\end{table}

\begin{equation}
\resizebox{0.92\linewidth}{!}{
$    \text{DOA}_{k} = \frac{1}{Z} \sum_{a, b \in S} \delta\left(\bm{Mas}_{s_a, c_k}, \bm{Mas}_{s_b, c_k}\right) \cdot 
    \frac{
    \sum_{j=1}^{M} \bm{Q}_{j,k} \land \varphi(j, a, b) \land \delta\left(r_{a j}, r_{b j}\right)
    }{
        \sum_{j=1}^{M} \bm{Q}_{j,k} \land \varphi(j, a, b) \land \mathbb{I}\left(r_{a j} \neq r_{b j}\right)
    },$
}
\end{equation}
where $Z=\sum_{a, b \in S}\delta(\bm{Mas}_{s_a, c_k}, \bm{Mas}_{s_b, c_k})$, $\bm{Q}_{j,k}=1$ indicates that exercise $e_j$ is related to concept $c_k$, $\varphi(j, a, b)$ determines whether both students $s_a$ and $s_b$ answered $e_j$, $r_{aj}$ is the response of $s_a$ to $e_j$, and $\mathbb{I}\left(r_{a j} \neq r_{b j}\right)$ determines whether their responses are different. $\delta(r_{a j}, r_{b j})$ is $1$ for a correct response by $s_a$ and an incorrect response by $s_b$, and $0$ otherwise.

\subsection{Effectiveness Analysis of Embedding Enhancement in CD scenarios and CAT}\label{apx:tab:CD&CAT}

\textbf{The performance of EduEmbed in CD scenarios and CAT.}

$\bullet$ \textbf{Transductive CD.} In transductive CD, textual embeddings offer limited benefits and can even underperform ID embeddings, as generalization demands are low and ID embeddings are well-optimized with encoders such as graph neural networks. Since textual embeddings are not further tuned during representation learning, they involve fewer trainable parameters and therefore underperform. However, EduEmbed integrates the ID paradigm to secure a strong lower bound and maintain competitiveness.

$\bullet$ \textbf{Inductive CD.} In inductive CD, textual embeddings yield notable gains by encoding richer information than sparse handcrafted features used in IDCD. Yet, these sparse features retain an interpretability advantage, as their structured patterns are more transparent than dense textual representations.

$\bullet$ \textbf{Zero-shot CD.} Textual semantics yield substantial gains in zero-shot CD across cross-domain and cross-subject settings~\cite{LRCD}. LRCD, which fully relies on semantic features, markedly outperforms methods with limited or no semantic use (e.g., TechCD, ZeroCD). Building on this, EduEmbed fine-tunes LMs to align with CD objectives, further bridging the gap and enhancing zero-shot performance.

$\bullet$ \textbf{CAT.} In CAT, textual semantics enhance performance at all stages, with the greatest gains in early phases when ID embeddings are weak and generalize poorly. As testing progresses and ID embeddings become refined, the setting converges toward transductive CD, where ID-based methods regain superiority.

\textbf{The Performance of Zero-shot CD with Overlapping Students.} We construct a new dataset, SLP$^*$, where the source and target domains share overlapping students. This dataset contains 312 students, 882 exercises, and 38 knowledge concepts, with a total of 32,996 response logs. We set SLP-CHI as the source domain and SLP-Math as the target domain. The experimental results are shown in Table~\ref{tab:Zero_overlap}, where EduEmbed consistently demonstrates strong performance compared to other methods.

\subsection{Ablation Study}\label{apx:ablation}
\begin{table}[!t]
\centering
\captionsetup{type=table}
\caption{Ablation study in transductive CD.}
\resizebox{0.7\linewidth}{!}{
\begin{tabular}{c|ccc}
  \toprule
  \textbf{Metric} & \textbf{\makecell{EduEmbed\\w/o IDI}} & \textbf{\makecell{EduEmbed\\w/o IDC}} & \textbf{EduEmbed} \\
  \midrule
  \textbf{AUC} & 82.21 & 82.05 & \textbf{82.23} \\
  \textbf{ACC} & 74.40 & 74.27 & \textbf{74.45} \\
  \textbf{DOA} & 77.50 & 77.59 & \textbf{77.85} \\
  \bottomrule
\end{tabular}
}
\label{tab:ablation_trans}
\end{table}

\begin{table*}[!t]
\centering
\caption{Ablation study in CAT.}

\resizebox{0.8\linewidth}{!}{
\begin{tabular}{c|c|cccccc}
\toprule
\multicolumn{2}{c|}{\textbf{Metric}} & \multicolumn{6}{c}{\textbf{AUC / ACC (\%)}} \\
\midrule
\textbf{CD Model} & \textbf{Step} & \makecell{\textbf{EduEmbed}\\\textbf{w/o RaIF}} & \makecell{\textbf{EduEmbed}\\\textbf{w/o RsR}} & \makecell{\textbf{EduEmbed}\\\textbf{w/o TA}} & \makecell{\textbf{EduEmbed}\\\textbf{w/o IDI}} & \makecell{\textbf{EduEmbed}\\\textbf{w/o IDC}} & \textbf{EduEmbed} \\
\midrule
\multirow{3}[1]{*}{\textbf{IRT}}
& \textbf{5}  & 67.72 / 61.83 & 73.71 / 58.49 & 73.21 / 66.91 & 76.69 / 69.30 & 76.29 / 67.87 & \textbf{77.45} / \textbf{70.40} \\
& \textbf{10} & 73.91 / 65.54 & 76.00 / 69.53 & 73.78 / 67.34 & \textbf{79.15} / \textbf{71.67} & 78.20 / 70.64 & 79.02 / 71.48 \\
& \textbf{15} & 74.95 / 68.72 & 80.05 / 71.94 & 74.64 / 67.81 & 81.27 / \textbf{73.50} & 78.92 / 69.87 & \textbf{81.33} / 73.38 \\
\midrule
\multirow{3}[1]{*}{\textbf{NCD}}
& \textbf{5}  & 62.26 / 57.20 & 72.25 / 65.64 & 70.65 / 64.41 & 69.83 / 62.57 & \textbf{73.37} / 64.07 & 72.36 / \textbf{65.74} \\
& \textbf{10} & 65.30 / 61.62 & 75.69 / 67.12 & \textbf{77.35} / \textbf{69.80} & 75.92 / 69.31 & 73.48 / 68.05 & 77.26 / 69.57 \\
& \textbf{15} & 66.93 / 63.08 & 76.54 / 67.88 & \textbf{78.78} / \textbf{71.55} & 78.17 / 70.15 & 77.89 / 70.85 & 78.40 / 70.20 \\
\bottomrule
\end{tabular}
}
\label{tab:ablation_CAT}

\end{table*}

In this subsection, we provide the additional experimental results of ablation study in transductive CD and CAT in Table~\ref{tab:ablation_trans} and~\ref{tab:ablation_CAT}. 

\textbf{Settings.} As for zero-shot and inductive CD, ID embeddings of new entities provide no useful information. Therefore, EduEmbed does not integrate them in these settings. Accordingly, experiments on \emph{EduEmbed-w/o-IDI} and \emph{EduEmbed-w/o-IDC} are omitted for inductive, cross-domain, and cross-subject CD. For inductive CD and CAT, the MLPs in the text adapter are replaced with a linear layer in \emph{EduEmbed-w/o-TA} to satisfy the dimension transfer required by the interaction function.

\textbf{Detailed Analysis.} 

$\bullet$ \textbf{Transductive CD.} \emph{EduEmbed-w/o-RaIF} achieves strong AUC performance, suggesting limited gains since ID embeddings are already well-trained. Nevertheless, EduEmbed still offers clear advantages in both accuracy and interpretability, underscoring its effectiveness in cognitive modeling.

$\bullet$ \textbf{Inductive CD.} \emph{EduEmbed-w/o-TA} also performs well, likely because MLPs add parameters but risk overfitting. These results validate the textual adapter framework, showing that even a simple linear layer ensures robust performance and offering insights for future adapter design.

$\bullet$ \textbf{Zero-shot CD.} EduEmbed shows weaker interpretability than \emph{EduEmbed-w/o-RsR}, likely due to the lack of explicit semantics in role embeddings, which is a limitation more evident in cross-domain CD requiring semantic generalization. Still, its strong predictive accuracy highlights the effectiveness of the role embedding design.

$\bullet$ \textbf{CAT.} Similar to inductive CD, \emph{EduEmbed-w/o-TA} achieves reasonable performance in early CAT stages. In contrast, \emph{EduEmbed-w/o-IDI} and \emph{EduEmbed-w/o-IDC} underperform at step 5 due to immature ID embeddings introducing noise. As CAT progresses, ID embeddings strengthen, and EduEmbed exhibits clear gains at steps 10 and 15, demonstrating the effectiveness of RaIF, as proposed in Section~\ref{sec:RaIF}.

\begin{figure}[!t]
  \centering
\begin{minipage}{0.49\linewidth}\centering
    \includegraphics[width=0.99\textwidth]{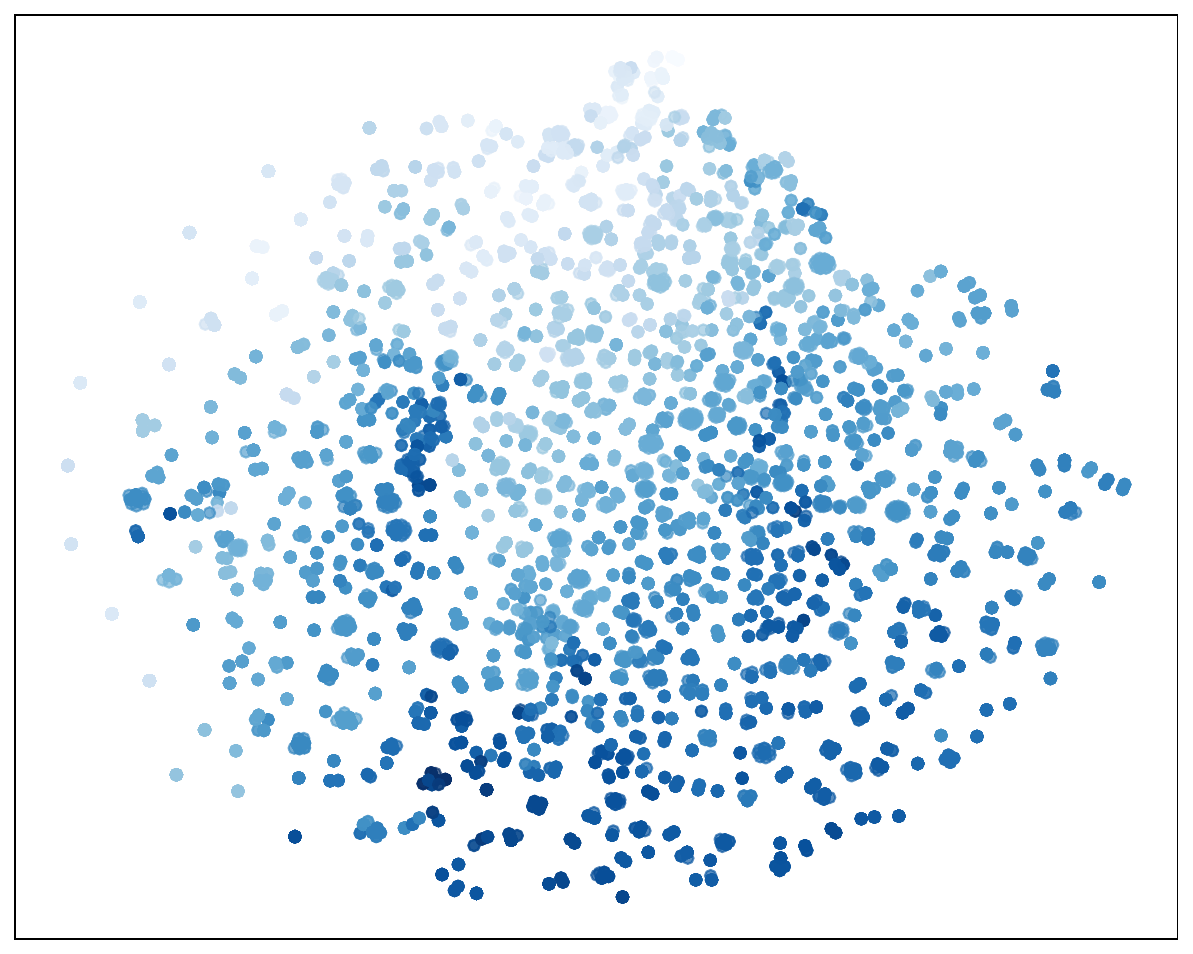}\\
    (a) EduEmbed-w/o-TA
\end{minipage}
\begin{minipage}{0.49\linewidth}\centering
    \includegraphics[width=0.99\textwidth]{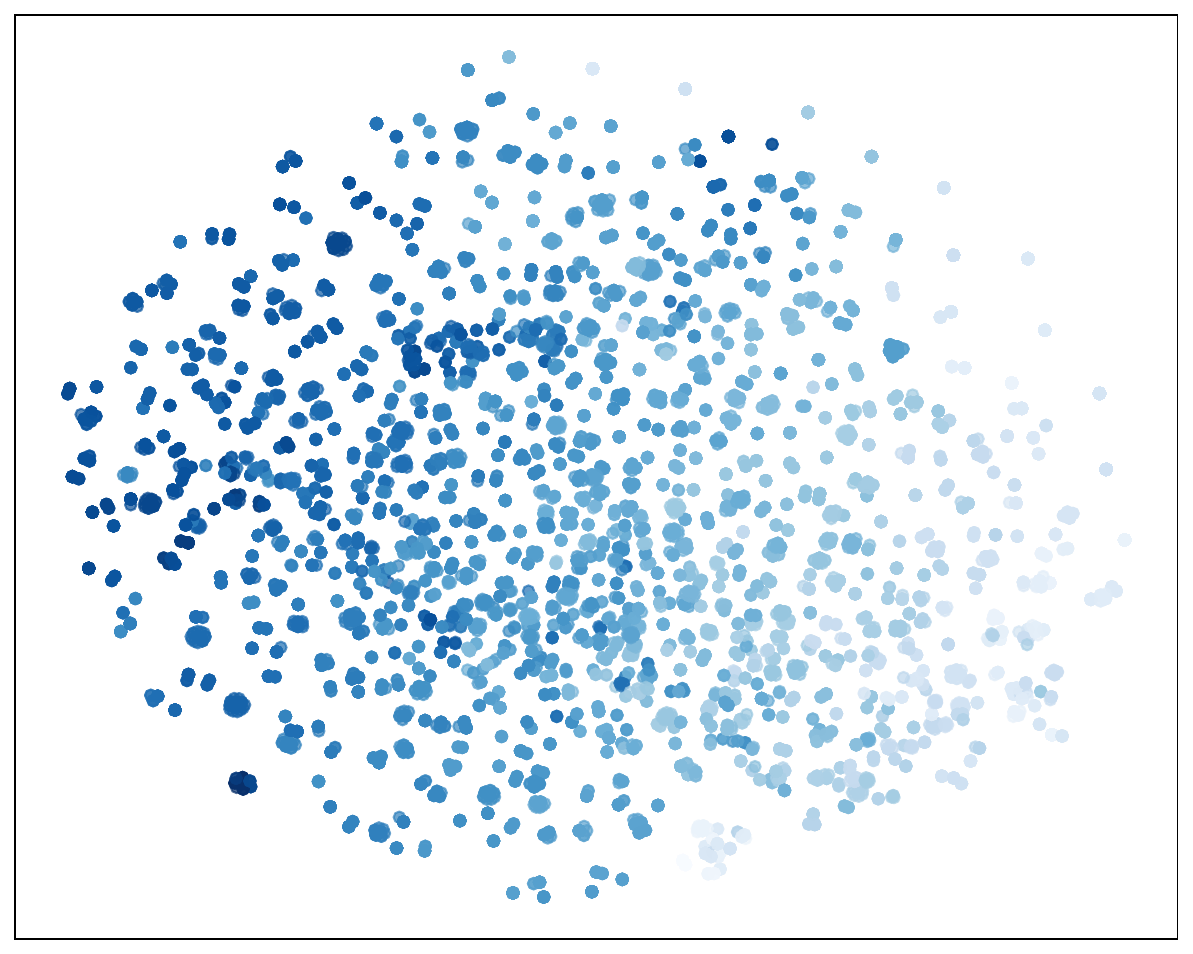}
    (b) EduEmbed
\end{minipage}
  \caption{Visualization of students' mastery levels on SLP-Math.}
  \label{fig:text_adapter}
\end{figure}

\textbf{Visualization of Mastery Levels.} To further evaluate the contribution of the Texual Adapter, we visualize students’ mastery levels on the SLP-Math dataset via t-SNE~\cite{Tsne}, with darker shades indicating higher correct rate. Using transductive CD as a case study, as shown in Figure~\ref{fig:text_adapter}, EduEmbed demonstrates clearer clustering and smoother progression, underscoring the interpretability benefits of the Textual Adapter.

The results of the ablation study indicate that designs in both RaIF and AaRI are crucial to the overall effectiveness of EduEmbed. 

\begin{figure}[htbp]
\centering
\begin{minipage}{0.49\linewidth}\centering
    \includegraphics[width=0.95\textwidth]{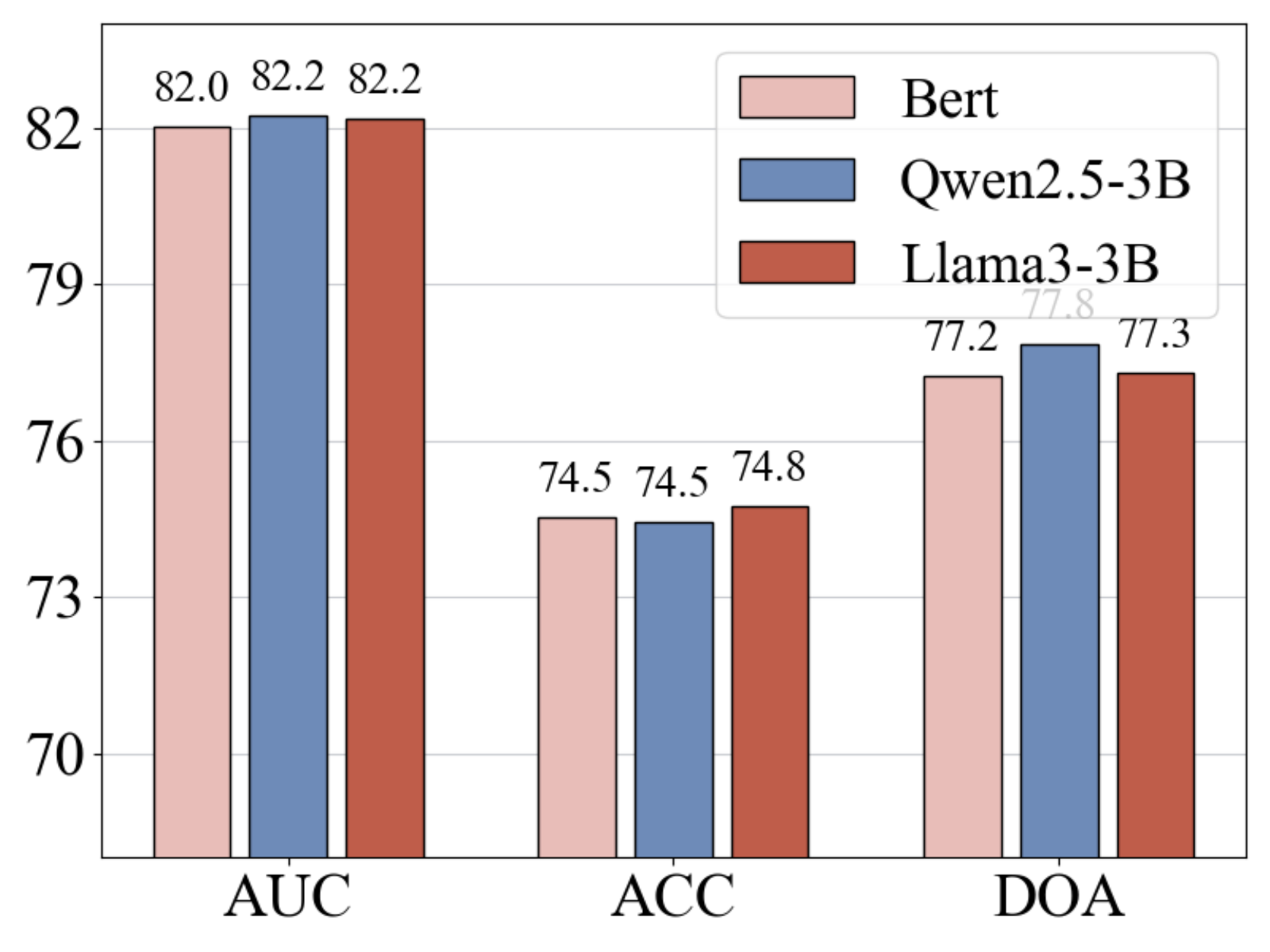}\\
    (a) Transductive CD
\end{minipage}
\begin{minipage}{0.49\linewidth}\centering
    \includegraphics[width=0.95\textwidth]{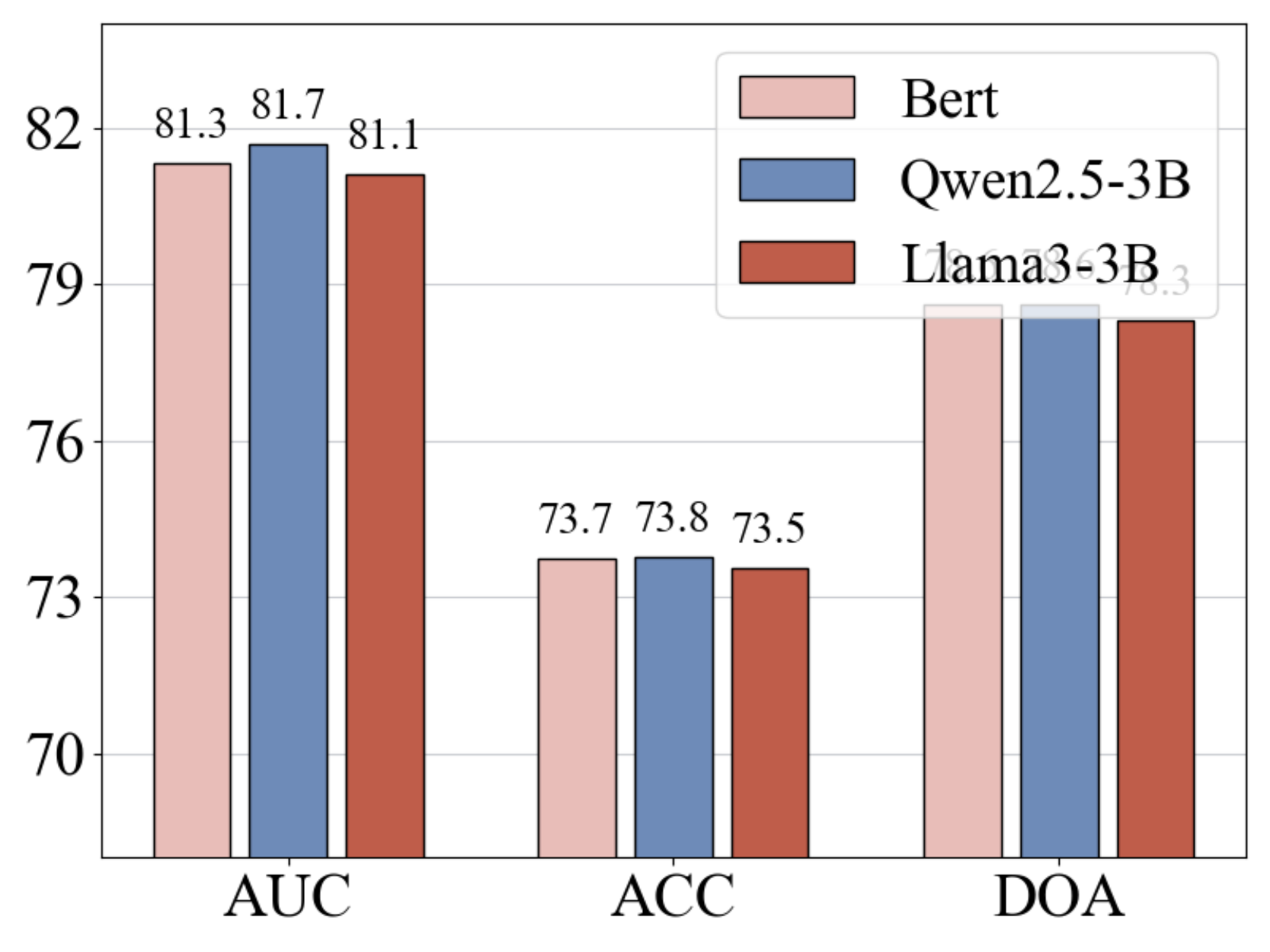}\\
    (b) Inductive CD
\end{minipage}
\\
\begin{minipage}{0.49\linewidth}\centering
    \includegraphics[width=0.95\textwidth]{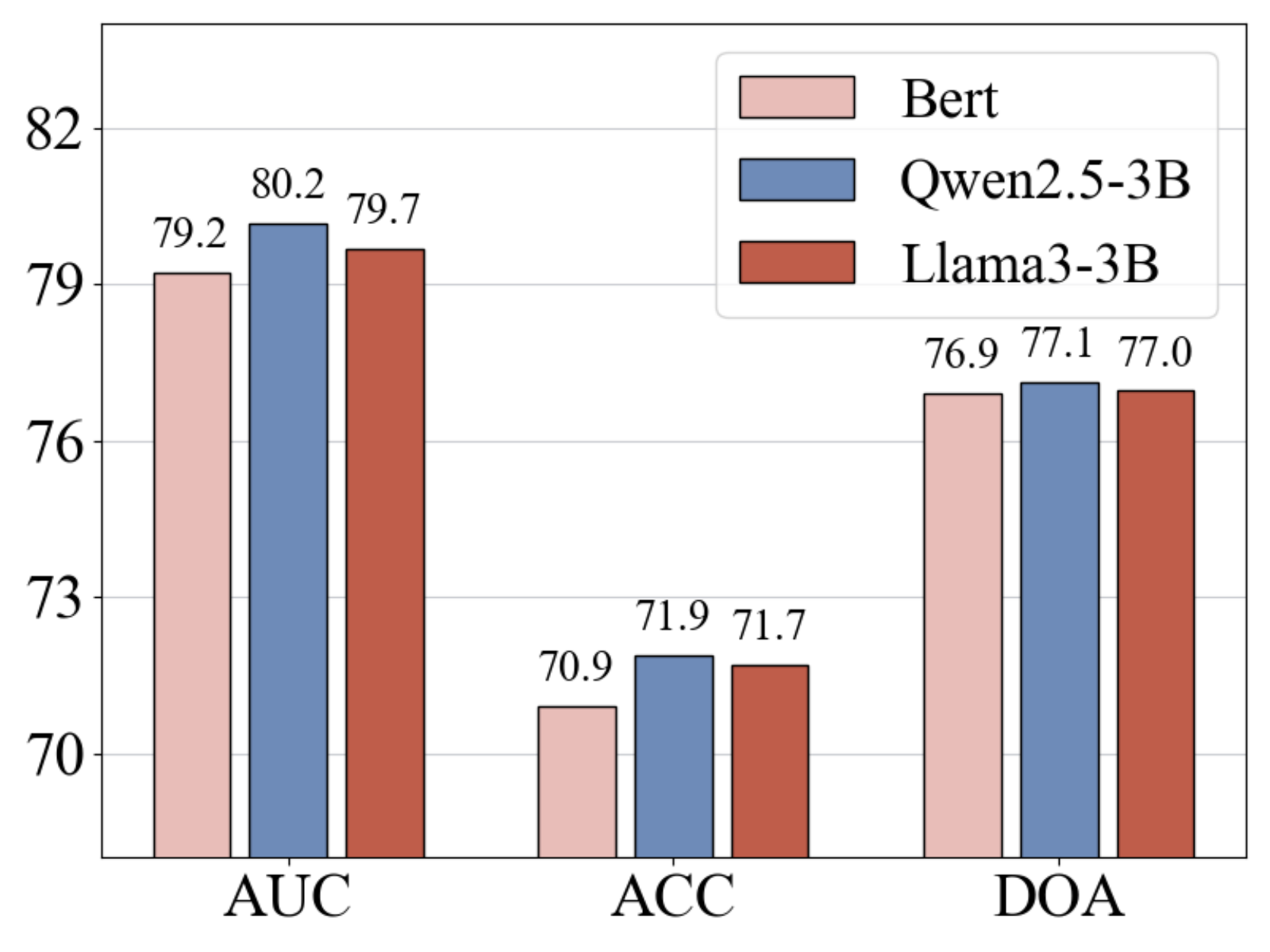}\\
    (c) Cross-Domain CD
\end{minipage}
\begin{minipage}{0.49\linewidth}\centering
    \includegraphics[width=0.95\textwidth]{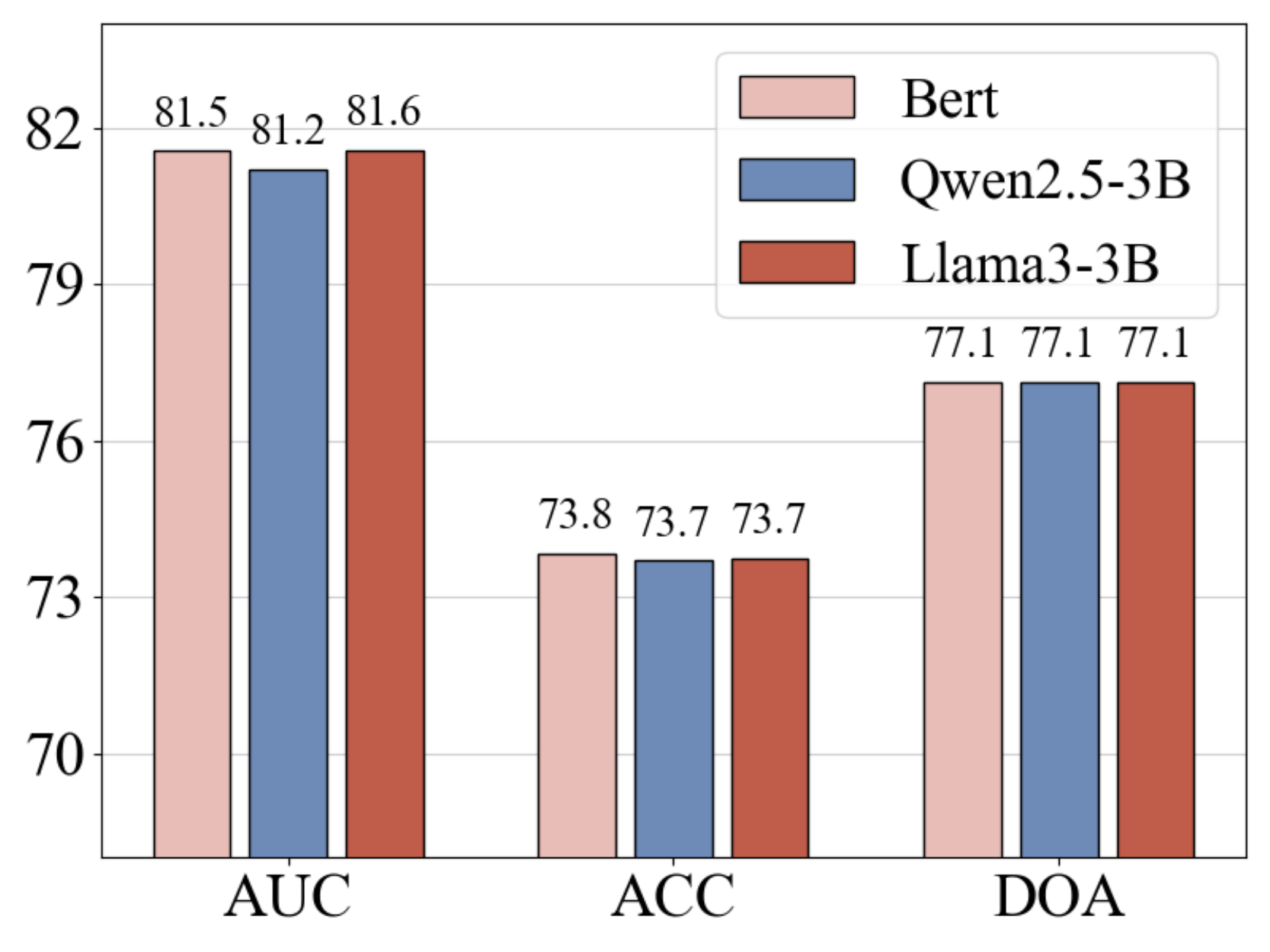}\\
    (d) Cross-Subject CD
\end{minipage}
\caption{Comparison of LMs types in four CD tasks.}
\label{fig:exp:type}
\end{figure}

\begin{figure}[htbp]
\centering
\begin{minipage}{0.49\linewidth}\centering
    \includegraphics[width=0.95\textwidth]{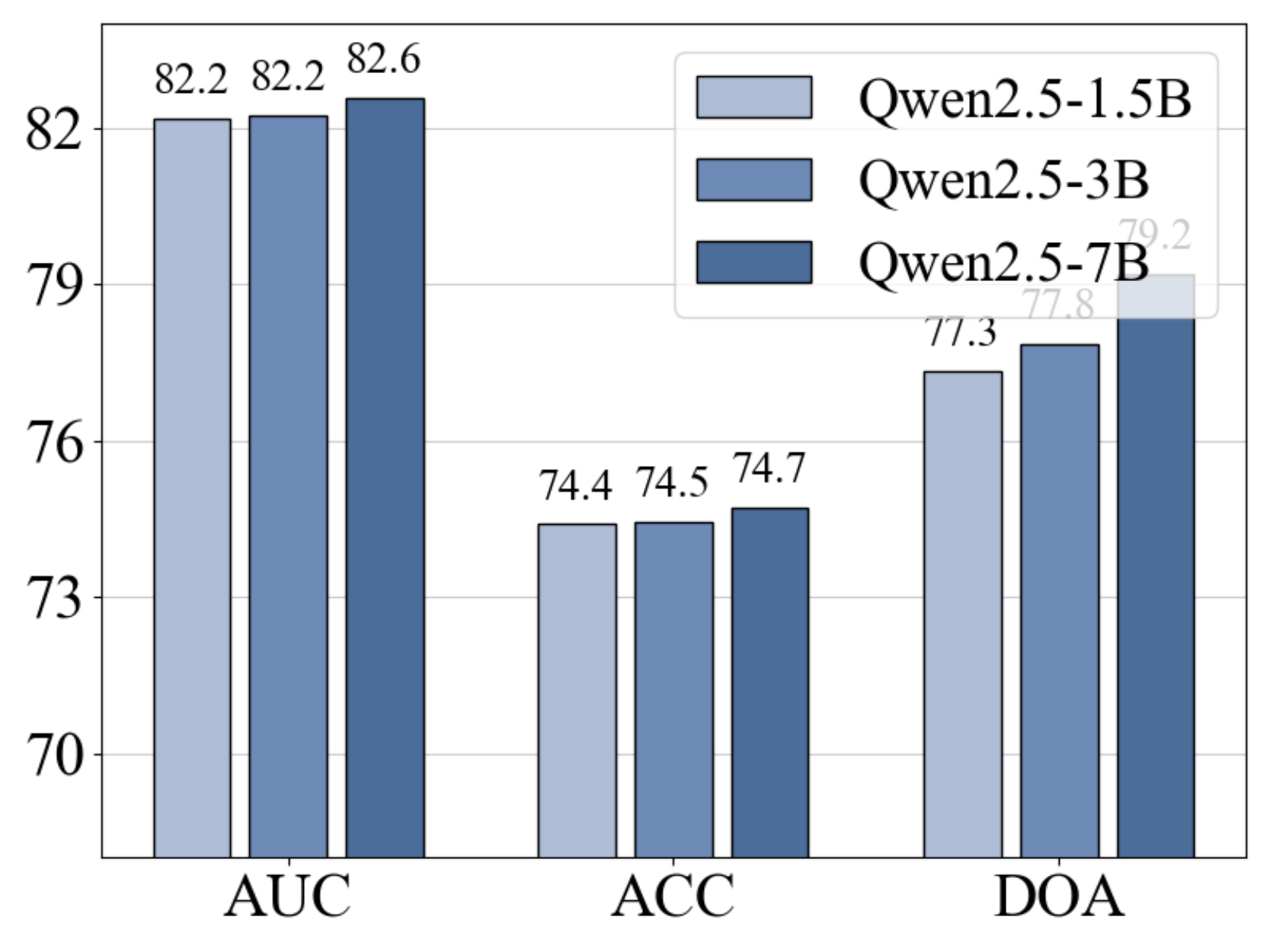}\\
    (a) Transductive CD
\end{minipage}
\begin{minipage}{0.49\linewidth}\centering
    \includegraphics[width=0.95\textwidth]{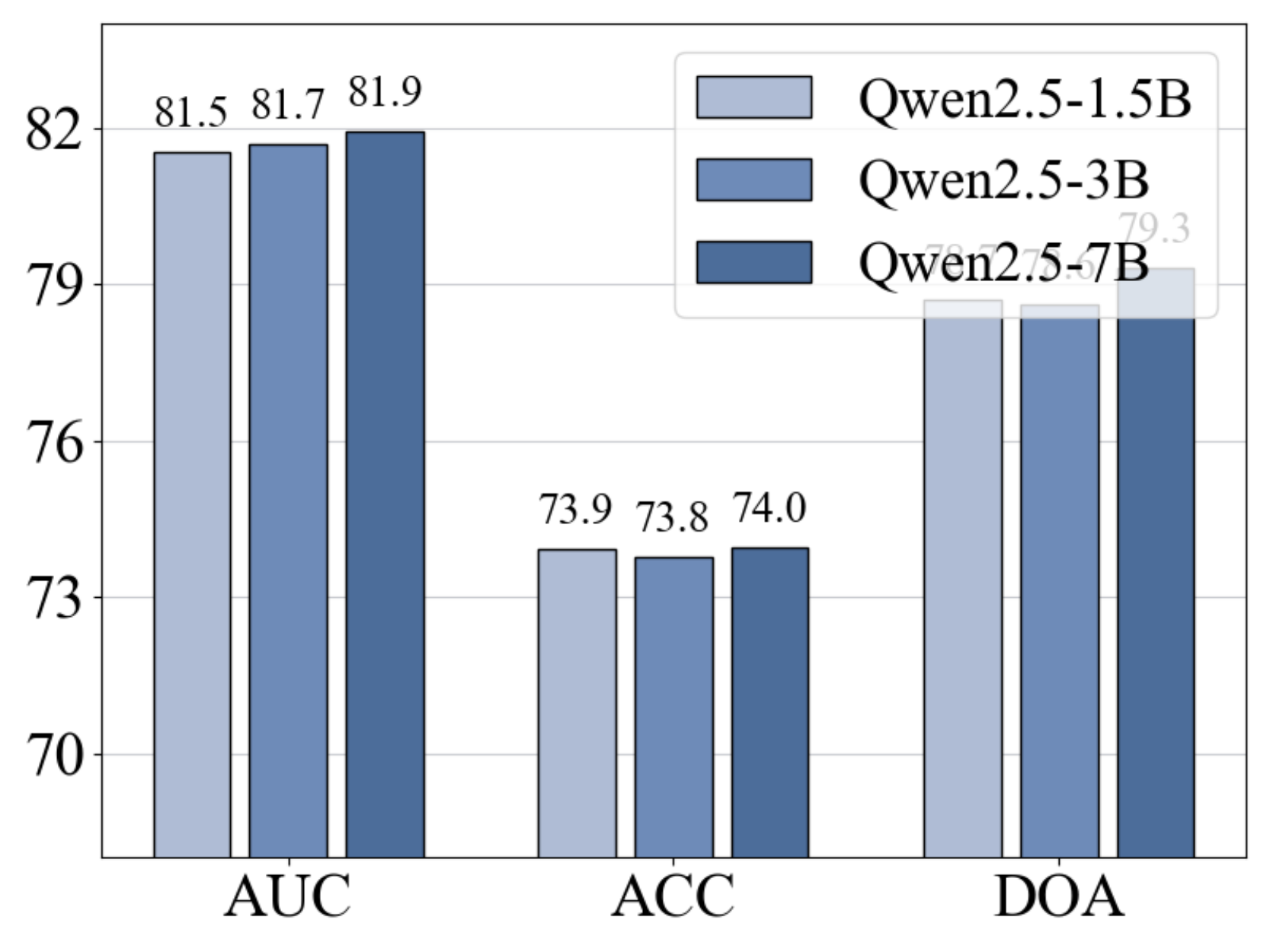}\\
    (b) Inductive CD
\end{minipage}
\\
\begin{minipage}{0.49\linewidth}\centering
    \includegraphics[width=0.95\textwidth]{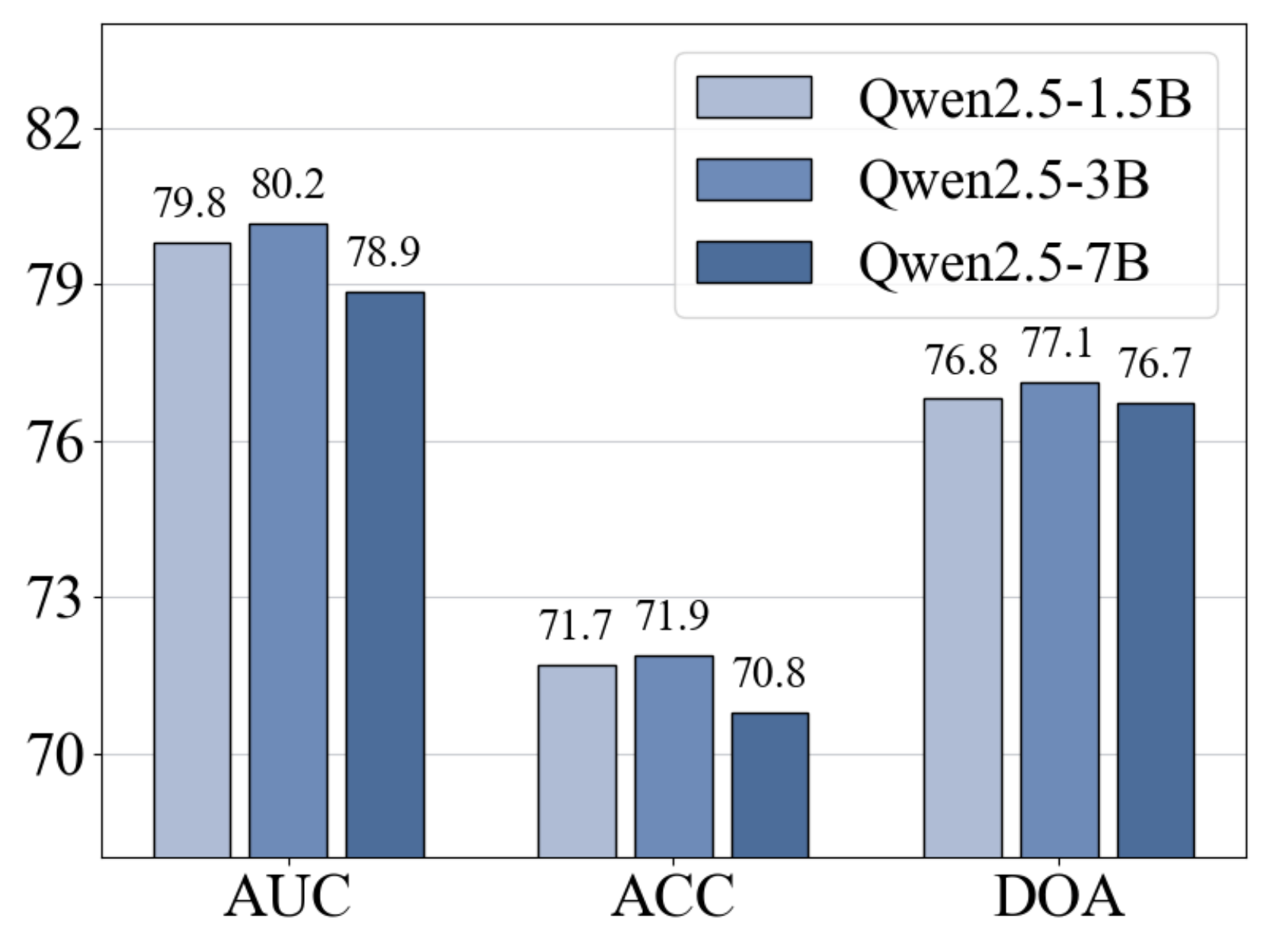}\\
    (c) Cross-Domain CD
\end{minipage}
\begin{minipage}{0.49\linewidth}\centering
    \includegraphics[width=0.95\textwidth]{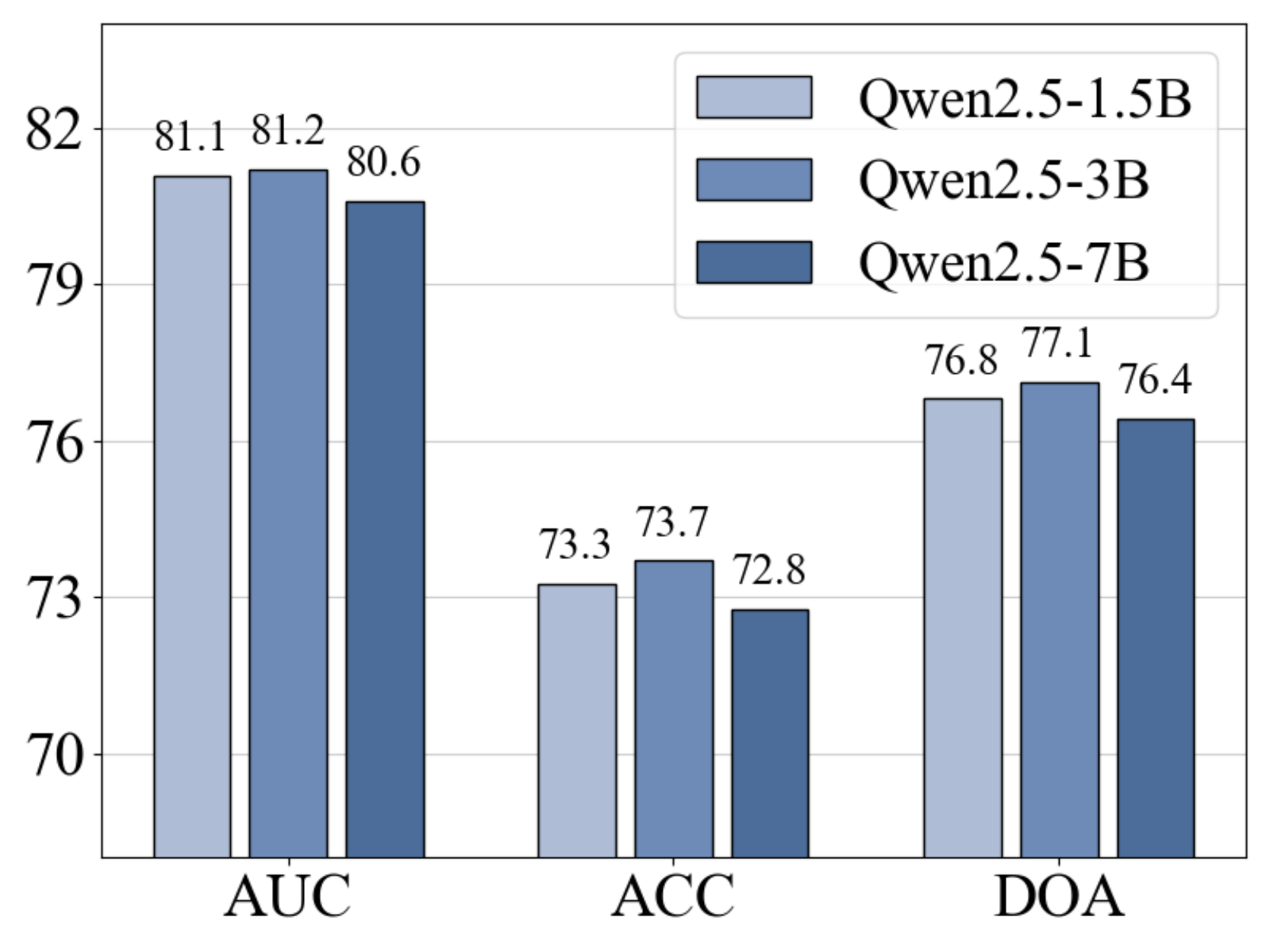}\\
    (d) Cross-Subject CD
\end{minipage}
\caption{Comparison of LMs scales in four CD tasks.}
\label{fig:exp:scale}
\end{figure}
\subsection{Details of LMs Scales and Types}\label{apx:scale&type}

In this subsection, we provide the performance results of EduEmbed with different types and scales of LMs in transductive CD, as shown in Figure~\ref{fig:exp:type} and~\ref{fig:exp:scale}.

\subsection{Text Selection Analysis}\label{apx:text_selection}
\begin{figure}[!t]
  \centering
\begin{minipage}{0.49\linewidth}\centering
    \includegraphics[width=0.99\textwidth]{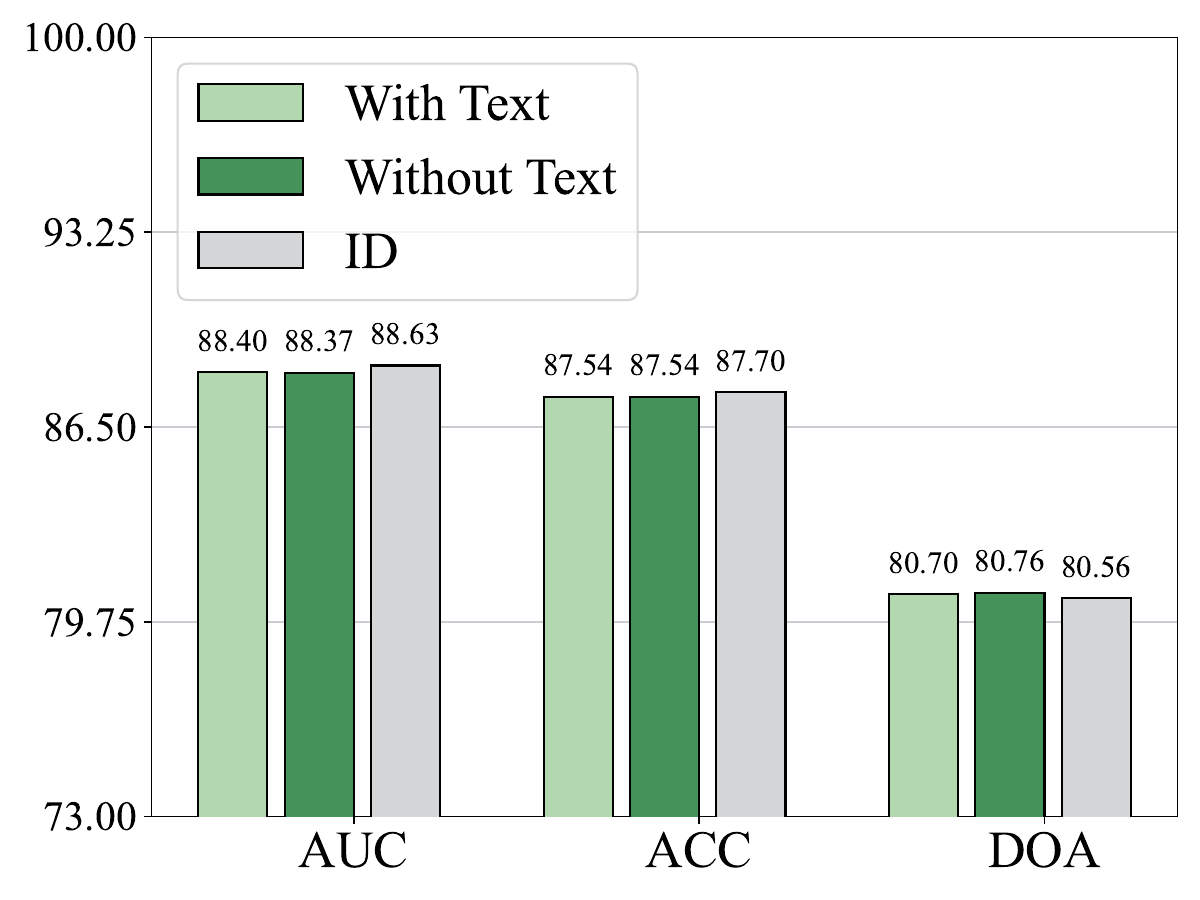}\\
    (a) Transductive CD
\end{minipage}
\begin{minipage}{0.49\linewidth}\centering
    \includegraphics[width=0.99\textwidth]{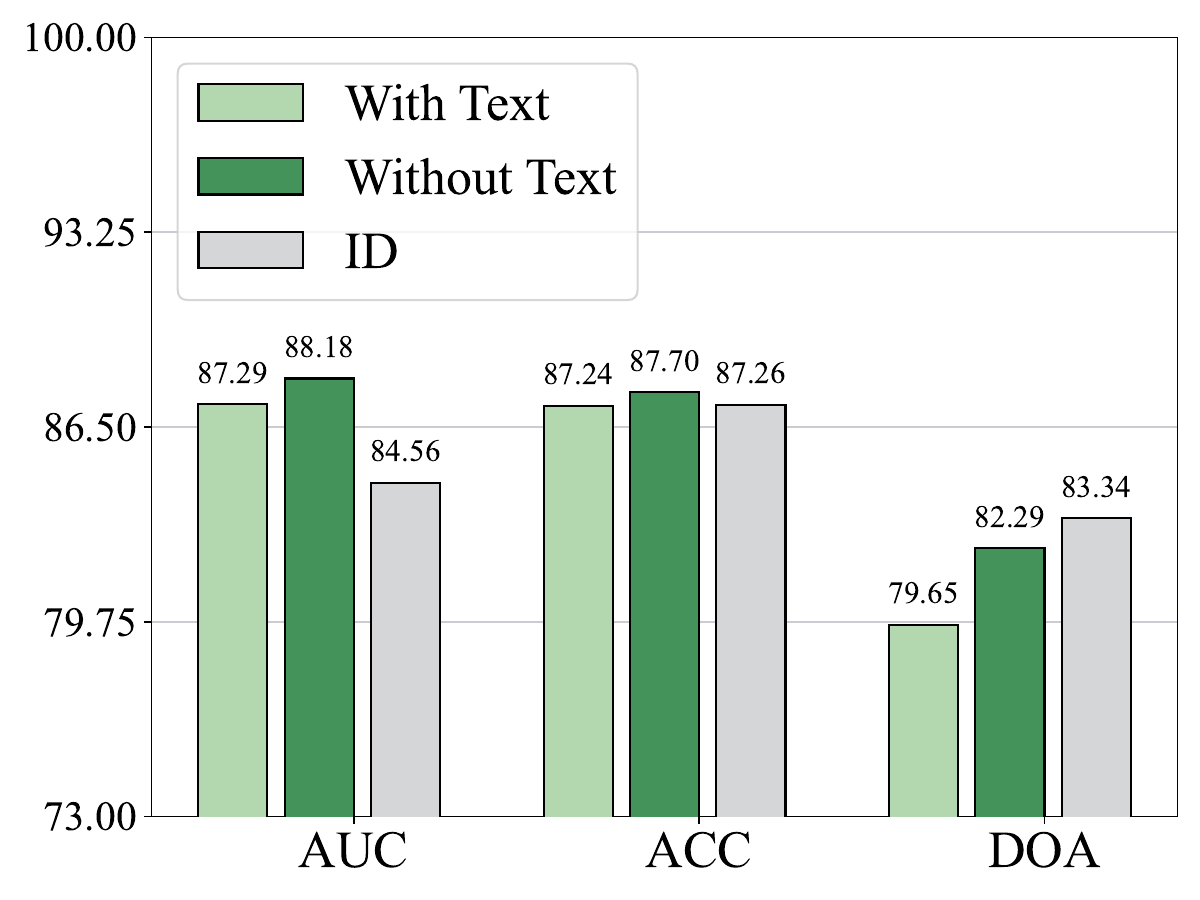}
    (b) Inductive CD
\end{minipage}
  \caption{Effect of text selection on MOOC. ``OL'' refer to baselines, specifically denoting ID embedding in transductive CD and IDCD for inductive CD.}
  \label{fig:text_select}
\end{figure}

In this subsection, we provide the details of text selection experiment. We extend the exercise attribute defined in Eq.~(\ref{eq:attr})  in Section~\ref{sec:role_specific} by incorporating textual content. Since the exercise content in MOOC is in Chinese, we adopt BERT-Base-Chinese~\cite{devlin2019bert} as the fine-tuned LM to ensure compatibility with the dataset. As shown in Figure~\ref{fig:text_select}, incorporating exercise content leads to modest performance fluctuations, likely due to the trade-off between added detail and potential noise of exercise content. This suggests that in datasets lacking exercise content, deriving attributes from response logs has minimal impact on model performance, especially when ultra-high prediction precision is unnecessary.

\begin{figure}[!t]
  \centering
\begin{minipage}{0.49\linewidth}\centering
    \includegraphics[width=0.99\textwidth]{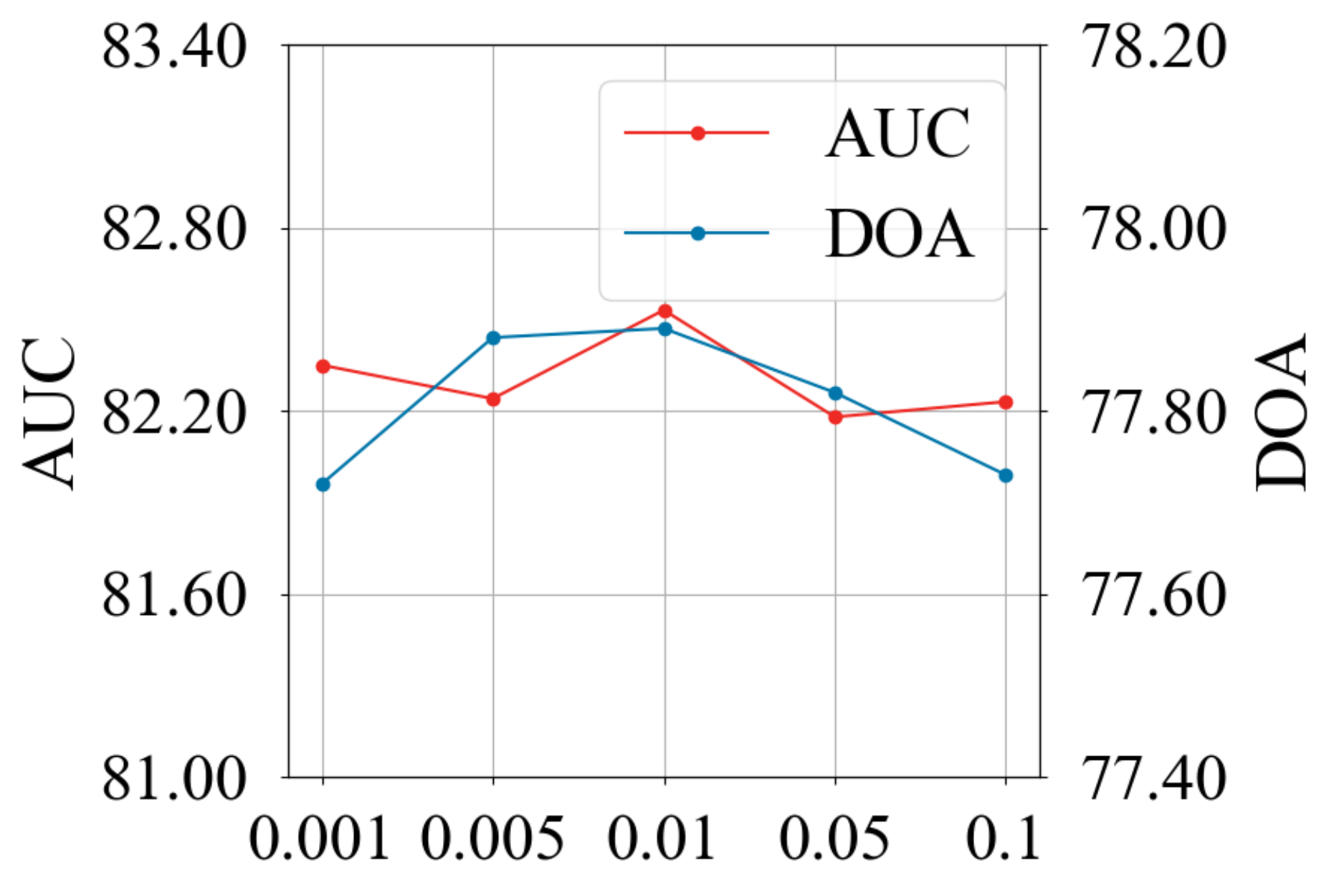}\\
    (a) Effect of $\alpha$
\end{minipage}
\begin{minipage}{0.49\linewidth}\centering
    \includegraphics[width=0.99\textwidth]{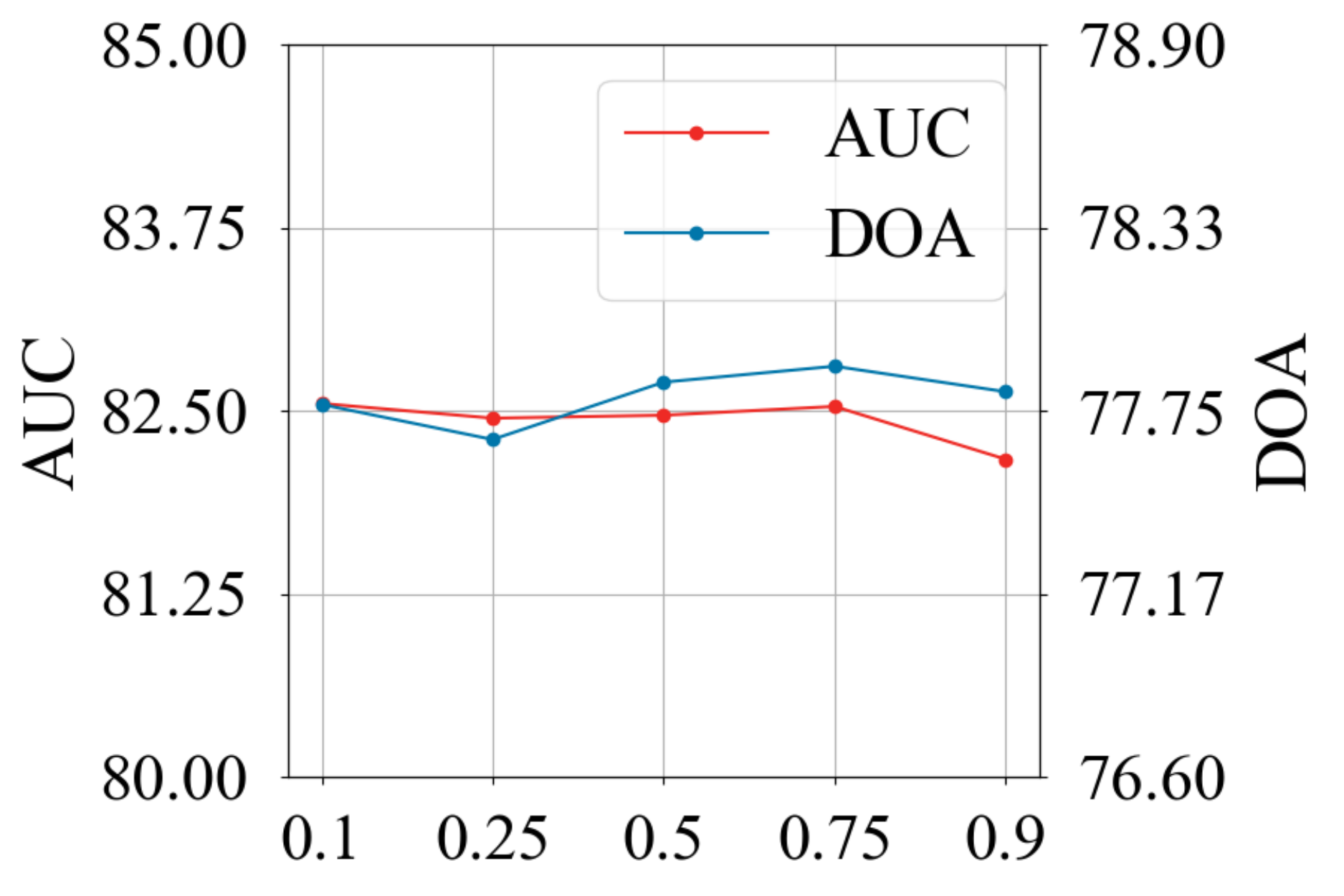}\\
    (b) Effect of $\lambda$
\end{minipage}
   \caption{Hyperparameter analysis on SLP-Math.}
  \label{fig:hyper_math}
\end{figure}

\subsection{Hyperparameter Analysis}\label{apx:hyper}

In this subsection, We present the performance of EduEmbed with different hyperparameter settings, as shown in Figure~\ref{fig:hyper_math}. We recommend setting $\alpha$ to 0.01 or 0.005 and $\lambda$ to 0.5 or 0.75 to generally yield relatively good performance in most cases.

\begin{table}[!h]
  \centering
  \caption{Comparison of EduEmbed and ``Text-Only'' on SLP-Math in transductive CD.}
    \begin{tabular}{c|c|c}
    \toprule
    Metric & Text-Only & EduEmbed \\
    \midrule
    AUC   & 75.53 & \textbf{82.23} \\
    ACC   & 68.93 & \textbf{74.45} \\
    DOA   & 76.60 & \textbf{77.85} \\
    \bottomrule
    \end{tabular}%
  \label{tab:disc_trans}%

\end{table}%
\section{Discussions} \label{apx:discuss}
\textbf{Performance Robustness in Low-Generalization Scenarios.} As discussed in Section~\ref{sec:emb_enhance} and Appendix~\ref{apx:tab:CD&CAT}, LMs show limitations in transductive CD compared to traditional ID-based models. By integrating ID information, EduEmbed ensures a reliable performance lower bound while flexibly adapting to various CD scenarios. Instead of pursuing a one-size-fits-all solution, EduEmbed is designed to flexibly adapt to various CD scenarios with minimal modification, highlighting its practical extensibility. As shown in Table~\ref{tab:disc_trans}, EduEmbed achieves superior performance on SLP-Math compared to the ``Text-Only'' variant using raw LM embeddings without fine-tuning, highlighting that direct use of textual features alone is suboptimal in transductive CD.


\textbf{Integration with Existing Learning Paradigms.} Given the effectiveness of mainstream ID embeddings in cognitive modeling, this work focuses on the fusion of textual embeddings with ID embeddings, to ensure EduEmbed’s compatibility across most CD tasks. Other paradigms, such as IDCD, which incorporate handcrafted interaction features as prior information, are also expected to be integrated. Notably, from a methodological perspective, EduEmbed is capable of being integrated with such paradigms. Exploring how textual embeddings can be effectively combined with increasingly diverse approaches remains an important direction for future research.

\textbf{Computational Cost.} Although fine-tuning LMs is generally time-consuming, our proposed decoupled EduEmbed mitigates this issue by freezing the textual embeddings by the LMs and applying them across different CD tasks. As a result, the fine-tuning process only needs to be conducted once, after which the representations can be stored locally. Therefore, in practical applications, the runtime of this component is virtually negligible, significantly improving the overall efficiency and usability of our framework.


\clearpage
\end{document}